\documentclass[journal]{IEEEtran}
\usepackage{float}
\usepackage{amssymb}
\usepackage{amsthm}
\usepackage{amsfonts}
\usepackage{mathrsfs}
\usepackage{graphicx}
\usepackage{amsmath,bm}
\usepackage{algorithmic}
\usepackage{algorithm}
\usepackage{setspace}
\usepackage{color}
\usepackage{subfig}
\usepackage{caption}
\usepackage{url}
\usepackage{booktabs}
\usepackage{multirow}
\usepackage{arydshln}

\begin{document}

\title{Margin-aware Fuzzy Rough Feature Selection: Bridging Uncertainty Characterization and Pattern Classification}

\author{Suping Xu, Lin Shang,~\IEEEmembership{Member,~IEEE,} Keyu Liu, \\
Hengrong Ju, Xibei Yang, and Witold Pedrycz,~\IEEEmembership{Life Fellow,~IEEE}

\thanks{S. Xu and W. Pedrycz are with the Department of Electrical and Computer Engineering, University of Alberta, Edmonton, AB T6G 2R3, Canada. (E-mail: supingxu@yahoo.com; suping2@ualberta.ca, wpedrycz@ualberta.ca)}
\thanks{L. Shang is with the Department of Computer Science and Technology, Nanjing University, Nanjing 210023, China, and also with the State Key Laboratory for Novel Software Technology, Nanjing University, Nanjing 210023, China. (E-mail: shanglin@nju.edu.cn)}
\thanks{K. Liu and X. Yang are with the School of Computer, Jiangsu University of Science and Technology, Zhenjiang 212003, China. (E-mail: kyliu@just.edu.cn, jsjxy\_yxb@just.edu.cn)}
\thanks{H. Ju is with the School of Artificial Intelligence and Computer Science, Nantong University, Nantong 226019, China. (E-mail: juhengrong@ntu.edu.cn)}

\thanks{Manuscript received X X, 2025; revised X X, 2025.}}

\markboth{IEEE TRANSACTIONS ON FUZZY SYSTEMS,~Vol.~X, No.~X, X~2025}%
{Shell \MakeLowercase{\textit{et al.}}: Bare Demo of IEEEtran.cls for IEEE Journals}

\maketitle

\begin{abstract}
Fuzzy rough feature selection (FRFS) is an effective means of addressing the curse of dimensionality in high-dimensional data. By removing redundant and irrelevant features, FRFS helps mitigate classifier overfitting, enhance generalization performance, and lessen computational overhead. However, most existing FRFS algorithms primarily focus on reducing uncertainty in pattern classification, neglecting that lower uncertainty does not necessarily result in improved classification performance, despite it commonly being regarded as a key indicator of feature selection effectiveness in the FRFS literature. To bridge uncertainty characterization and pattern classification, we propose a Margin-aware Fuzzy Rough Feature Selection (MAFRFS) framework that considers both the compactness and separation of label classes. MAFRFS effectively reduces uncertainty in pattern classification tasks, while guiding the feature selection towards more separable and discriminative label class structures. Extensive experiments on 15 public datasets demonstrate that MAFRFS is highly scalable and more effective than FRFS. The algorithms developed using MAFRFS outperform six state-of-the-art feature selection algorithms.

\end{abstract}


%
\IEEEpeerreviewmaketitle

\section{Introduction}
\label{sec:1}

\IEEEPARstart{W}{ith} the rapid advancement of data acquisition technologies and storage solutions, real-world data in various applications often appear in high-dimensional form, accompanied by a multitude of features. Some of these features are essential for learning processes, whereas others may be redundant or irrelevant. The presence of unnecessary features not only reduces the generalization performance of learning models but also increases computational overhead. Feature selection, guided by multiple evaluation criteria, provides an effective mechanism to eliminate irrelevant or redundant features. Its objective is to derive an optimal subset of original features while preserving critical learnable information. Alternatively, feature selection aims to produce an optimal ranking of original features based on their importance, efficiently maximizing learnable information. An important advantage of feature selection is its capacity to retain the original semantic interpretability of selected features. This capability significantly facilitates high-dimensional characterization through low-dimensional data analysis \cite{WrightJ2022} and supports enhanced data visualization and comprehension \cite{LaurensV2008,GisbrechtA2015}. Consequently, feature selection has received growing attention in recent years, especially in the fields of machine learning \cite{YangQ2025}, pattern recognition \cite{ZhaoY2025}, and data mining \cite{ZhangX2024}.

Rough set (RS) theory, a powerful methodology within granular computing, has emerged as an influential approach for managing imprecision, vagueness, and uncertainty in intelligent decision systems. Fuzzy rough set (FRS) theory \cite{DidierD1990} extends RS by integrating the strengths of RS and fuzzy set theories, particularly through fuzzy similarity relations. This integration enables finer information granulation and reduces information loss typically encountered during data discretization in RS. Due to the improved ability of fuzzy similarity relations to preserve distinctions among samples, fuzzy rough feature selection (FRFS) \cite{TsangE2008} demonstrates greater potential to improve feature subset performance compared to traditional RS-based feature selection. Consequently, numerous FRFS algorithms have been applied to diverse pattern
classification tasks in recent years. A comprehensive review of existing literature indicates that prevalent FRFS studies mainly fall into two fundamental categories: (i) \emph{developing measures to characterize the uncertainty of a task} and (ii) \emph{designing frameworks to search for qualified feature subsets}, which are discussed further in Section \ref{sec:2}. In the former, task uncertainty commonly reflects the amount of learnable information, such as classification information. Widely adopted uncertainty measures are derived from various perspectives, including belongingness \cite{JensenR2009,WangC2022,HuangZ2022,AnS2023}, degree of disorder \cite{ZhangX2016,ZhangX2020,WanJ2021,WangZ2023,WanJ2023,DaiJ2024,YangY2024}, and quality of information granules \cite{YangJ2020,XiaD2023}. The second category employs exhaustive or heuristic search strategies, forming feature selection frameworks suitable for practical requirements such as computational efficiency \cite{YangY2024,XiaD2023,DaiJ2018}, stability of features \cite{JiangZ2021}, or specific settings including online learning \cite{ZhangX2020,HuangW2023,ZhangC2025}, multi-label learning \cite{XuS2016,WangZ2024}, and weakly-supervised learning \cite{LiuK2023,ZhouN2025}.

It is worth noting that, for a given pattern classification task, although numerous FRS-based uncertainty measures and FRFS frameworks exist for selecting qualified feature subsets, the amount of learnable information--captured by these uncertainty measures--in a chosen subset may not consistently correlate closely with the classification performance of the subset as evaluated by widely-used supervised learners such as CART, SVM, and KNN. The reasons for this phenomenon are twofold. First, most FRFS algorithms belong to the ``filter'' type and are thus classifier-agnostic. This means that the selected subset--although appearing optimal according to uncertainty-based feature selection criteria--may not necessarily be optimal from the standpoint of subsequent classifiers, despite providing interpretable semantic insights.

Second, FRS-based uncertainty measures typically quantify the degree to which a feature space effectively characterizes, describes, or predicts the corresponding label space. However, they do so exclusively through the lens of various fuzzy similarity relations, commonly constructed using distance measures \cite{WangC2019}, correlation coefficients \cite{YuD2007}, fuzzy logic operators \cite{YeJ2021}, kernel functions \cite{HuQ2011,HuM2024}, among others. Consequently, it is intuitive that feature subsets selected based on these uncertainty measures are more likely to deliver significant improvements in classification performance, primarily when employed in conjunction with fuzzy rough classifiers \cite{HuQ2011FSS,ZhaoS2015,VluymansS2016} that adopt the same fuzzy similarity relations consistently during both the learning and prediction phases. Thus, a critical question naturally arises: How can we develop an effective and interpretable FRFS framework capable of identifying feature subsets that consistently exhibit superior predictive performance across diverse classifiers?

It is widely acknowledged that the effectiveness of pattern classification primarily depends on two key factors: the learning capability of the classifier itself and the intrinsic difficulty of the classification task. On one hand, classifiers with stronger representational capabilities can effectively capture subtle distinctions among various label classes, thus achieving better classification performance (although this aspect is not the focus of this paper). On the other hand, reducing the inherent difficulty of classification through data-driven methods is more universal and widely pursued. Specifically, enhancing compactness among samples within the same label class (reducing the within-class margin) helps classifiers more accurately identify samples belonging to that class. Meanwhile, expanding separation between samples from different label classes (increasing the between-class margin) creates a larger ``safe area'' around the classification decision boundary, thereby improving the classifier's generalization ability on unseen samples. Both of them often contribute significantly to improved classification outcomes. For example, LDA aims to identify an optimal projection direction by maximizing between-class variance while minimizing within-class variance in the projected space. Similarly, embedding techniques such as LLE and t-SNE transform high-dimensional data into low-dimensional spaces that enhance within-class compactness and between-class separation. These algorithms share a fundamental principle: Optimizing sample distribution in feature space by reducing within-class margins and simultaneously increasing between-class margins simplifies the pattern classification task. Unfortunately, although the algorithms mentioned above improve discrimination between label classes, the resulting feature space often lacks the clear semantic interpretability provided by feature selection algorithms.

In this paper, we propose a \textbf{M}argin-\textbf{a}ware \textbf{F}uzzy \textbf{R}ough \textbf{F}eature \textbf{S}election (MAFRFS) framework. MAFRFS employs a forward greedy search strategy (addition structure), which is widely adopted by various FRFS algorithms. At each iteration, MAFRFS not only seeks to effectively reduce uncertainty in pattern classification tasks, but also expects to achieve clearer and more discriminative label class boundaries. Specifically, we first construct a pool of preferred candidate features by selecting those features that can most significantly reduce classification uncertainty. To further assess the contributions of candidate features within this pool, we simultaneously consider compactness within the same label class and separation between different label classes, and introduce the concept of within-class margin, along with two types of between-class margins defined respectively by global and local strategies. We then formulate an objective aimed at explicitly minimizing the ratio of the within-class margin to between-class margin, thereby guiding the feature selection towards more separable and discriminative label class structures. Extensive experiments conducted on $15$ publicly available datasets demonstrate that MAFRFS outperforms FRFS, and further confirm that incorporating the uncertainty measure into MAFRFS leads to superior performance compared with six state-of-the-art feature selection algorithms.

In summary, our contributions are as follows.
\begin{itemize}
    \item By defining the within-class and between-class margins of label classes, MAFRFS aims to select features that enhance class separability and discriminability.   
    \item MAFRFS is highly scalable, capable of integrating with both existing and emerging uncertainty measures, and sufficiently adaptable to adjust its feature search strategy to meet practical requirements.
    \item Feature selection algorithms designed with MAFRFS demonstrate superior performance across $15$ pattern classification tasks.  
\end{itemize}

\section{Related Works}
\label{sec:2}
To date, a substantial body of studies on FRFS has been conducted. Broadly speaking, existing studies can be categorized into two fundamental groups: uncertainty measure development and framework design.

Existing measures for uncertainty characterization include fuzzy dependency \cite{JensenR2009,WangC2022,HuangZ2022}, fuzzy positive region \cite{AnS2023}, fuzzy information entropy \cite{ZhangX2016,ZhangX2020,DaiJ2024,YangY2024}, fuzzy mutual information \cite{WanJ2023,WangZ2023}, fuzzy information gain \cite{WanJ2021}, and fuzzy information granularity \cite{YangJ2020,XiaD2023}. The first two measures capture the belongingness of each sample to label classes, whereas entropy-based measures quantify disorder in pattern classification tasks. Fuzzy information granularity evaluates the quality and effectiveness of information granules. To improve classification performance achieved by selected feature subsets, several uncertainty measures have aimed at increasing discrimination between samples or reducing the adverse effects of noise. For instance, when analyzing high-dimensional data, the conjunction of multiple fuzzy similarity relations may diminish discriminative information in constructing FRS models. To tackle this, Wang et al. introduced a fuzzy distance \cite{WangC2019} for directly assessing fuzzy similarity relations between samples. Recognizing that samples from different label classes typically exhibit notable distinctions, Wang et al. proposed a label-class-aware FRS model \cite{WangC2017}, ensuring each sample achieves maximal membership in its respective label class. Addressing the sensitivity of traditional FRS models to noisy data, Li et al. designed a robust DC\_ratio model \cite{LiY2017} by defining the concept of different classes' ratio, which identifies noisy samples through neighborhood analysis and subsequently removes them, thereby enhancing robustness. Moreover, Jensen and Shen noted that the majority of FRFS algorithms might yield only near-minimal feature subsets, although these algorithms remain effective in substantially reducing feature dimensionality. Consequently, they proposed a fuzzy discernibility matrix \cite{JensenR2009} to identify distinguishing features among different samples, alongside a fuzzy discernibility function \cite{JensenR2009} evaluating how distinctly each sample is differentiated from others. Dai et al. further approached FRFS from a ``sample pair'' perspective, specifically considering only those sample pairs not distinguished by previously selected features during iteration. Their algorithms, RMDPS and WRMDPS, based on maximal discernibility pairs \cite{DaiJ2018}, were designed explicitly to enhance the time efficiency of FRFS.

Leveraging the aforementioned uncertainty measures, numerous feature selection frameworks have been proposed to effectively identify qualified feature subsets. Among these frameworks, exhaustive search (also known as brute force search) has the advantage of identifying all possible qualified feature subsets. However, due to its computationally intensive nature, exhaustive search faces significant challenges when applied to large-scale, real-world pattern classification tasks. Within FRFS, two commonly utilized exhaustive search frameworks are the discernibility matrix \cite{JensenR2009,SkowronA1992} and backtracking \cite{YangX2013} methods. In practice, heuristic searches—such as forward or backward greedy algorithms \cite{AnS2023,JiangZ2021,XuS2016,LiuK2023}—and various meta-heuristic search algorithms \cite{SunL2023,LuoC2023} have become increasingly prevalent, as they effectively balance computational efficiency and solution effectiveness, particularly when only one near-optimal qualified feature subset is required.

In these heuristic and meta-heuristic FRFS frameworks, the fitness function designed based on certain uncertainty measures is an essential component. It quantitatively evaluates the significance of candidate features in each iteration and continuously updates the selected subset by either adding the most informative feature or removing the least informative one. Additionally, various FRFS frameworks have been developed to cater specifically to different application scenarios. For instance, Zhang et al. \cite{ZhangX2020} and Huang et al. \cite{HuangW2023} respectively presented the incremental frameworks AIFWAR and IFS; AIFWAR addresses the issue of continuously arriving samples, while IFS is designed to handle scenarios where new features are sequentially introduced. In multi-label learning, Xu et al. \cite{XuS2016} proposed two algorithms, FRS-LIFT and FRS-SS-LIFT, guided by a FD function, aiming to identify discriminative label-specific features. Furthermore, considering that manually labeling large datasets can be laborious and time-consuming, Liu et al. \cite{LiuK2023} introduced SemiFREE, a weakly-supervised feature selection framework that maximizes feature relevance while simultaneously minimizing feature redundancy.

\section{Preliminaries}
A pattern classification task is described as a 3-tuple $<U,F,L>$, where $U=\{x_1,x_2,\ldots,x_n\}$ is a non-empty finite set of $n$ samples known as the universe of discourse, $F=\{f_1,f_2,\ldots,f_m\}$ is a set of $m$ features characterizing the samples, and $L$ denotes the label. $\mathscr{F}:U \rightarrow [0,1]$ is a fuzzy set \cite{DidierD1990} on $U$. For $\forall x_i \in U$, $\mathscr{F}(x_i) \in [0,1]$ is the membership degree of $x_i$ to $\mathscr{F}$. $\mathscr{F}(U)$ is the fuzzy power set of $U$. A 2-tuple $(U,\mathscr{R})$ is called a fuzzy approximation space, where $\mathscr{R} \in \mathscr{F}(U \times U)$ is a fuzzy binary relation on $U$, and $\mathscr{R}(x_i,x_j) \in [0,1]$ measures the relationship between samples $x_i$ and $x_j$.

A fuzzy binary relation $\mathscr{R}$ qualifies as a fuzzy $T$-similarity relation if it is \emph{reflexive}, \emph{symmetric} and \emph{$T$-transitive}. In this work, the $T$-norm is defined using the standard $min$ operator, and thus, a fuzzy $T$-similarity relation is simply referred to as a fuzzy similarity relation.

Let $\mathscr{R}_{F}$ denote the fuzzy similarity relation induced by feature $F$ on $U$, then the fuzzy similarity class $[x_i]^{fsim}_{F}$, associated with $x_i \in U$ under $\mathscr{R}_{F}$, is represented as:
\begin{align}
[x_i]^{fsim}_{F} = & \frac{\mathscr{R}_{F}(x_i,x_1)}{x_1}+\frac{\mathscr{R}_{F}(x_i,x_2)}{x_2}+\dots \nonumber\\
& +\frac{\mathscr{R}_{F}(x_i,x_j)}{x_j}+\dots+\frac{\mathscr{R}_{F}(x_i,x_n)}{x_n}.
\label{eq:1}
\end{align}
Here, ``$+$'' stands for an ``union'', and ``/'' acts as a separator.

For each $x_{i} \in U$, its label $L(x_i)\in \{l_1,l_2,\ldots,l_p\}$ is single and symbolic. Based on the label $L$, we can define an equivalence relation on $U$ as follows:
\begin{equation}
IND(L) = \{(x_i,x_j) \in U \times U: L(x_i)=L(x_j)\},
\label{eq:2}
\end{equation}
where $L$ partitions $U$ into $p$ boolean label classes (equivalence classes) denoted by $U/IND(L)=\{LC_1,LC_2,\ldots,LC_p\}$. Each $LC_q $ contains all samples assigned the label $l_q$, for $q = 1,2,\ldots,p$.

Given $<U,F,L>$, the fuzzy label of $x_i \in U$ is defined as:
\begin{equation}
FL(x_i) = \{\mathscr{L}_1(x_i),\mathscr{L}_2(x_i),\ldots,\mathscr{L}_p(x_i)\},
\label{eq:3}
\end{equation}
where $\mathscr{L}_q(x_i)$, for $q = 1,2,\ldots,p$, denotes the membership degree of $x_i$ to $LC_q \in U/IND(L)$, and is computed as follows \cite{WangC2017}:
\begin{equation}
\mathscr{L}_q(x_i) = \frac{| [x_i]^{fsim}_{F} \cap LC_q|}{| [x_i]^{fsim}_{F} |},
\label{eq:4}
\end{equation}
where $|[x_i]^{fsim}_{F}| = \sum_{x_j \in U}\mathscr{R}_{F}(x_i,x_j)$ is the fuzzy cardinality of $[x_i]^{fsim}_{F}$.

\subsection{Fuzzy Rough Set and Fuzzy Dependency}
\label{sec:3.1}

Given $<U,F,L>$, $LC_q$ is a label class. The fuzzy lower and upper approximations of $LC_q$ in $(U,\mathscr{R}_{F})$ are denoted as $\underline{\mathscr{R}_{F}}(LC_q)$ and $\overline{\mathscr{R}_{F}}(LC_q)$, respectively. For each $x_i \in U$, the membership degrees that $x_i$ belongs to $\underline{\mathscr{R}_{F}}(LC_q)$ and $\overline{\mathscr{R}_{F}}(LC_q)$ are defined as:
\begin{align}
\underline{\mathscr{R}_{F}}(LC_q)(x_i) & = \inf_{x_j \in U} S\big(N(\mathscr{R}_{F}(x_i,x_j)), \mathscr{L}_q(x_j)\big);\\
\overline{\mathscr{R}_{F}}(LC_q)(x_i) & = \sup_{x_j \in U} T\big(\mathscr{R}_{F}(x_i,x_j), \mathscr{L}_q(x_j)\big),
\label{eq:6}
\end{align}
where $T$ is a $T$-norm, $S$ is a $t$-conorm, and $N$ is a negation.

The pair $[\underline{\mathscr{R}_{F}}(LC_q),\overline{\mathscr{R}_{F}}(LC_q)]$ forms a fuzzy rough set of $LC_q$, with $\underline{\mathscr{R}_{F}}(LC_q)(x_i)$ and $\overline{\mathscr{R}_{F}}(LC_q)(x_i)$ indicating the degrees to which $x_i$ certainly and possibly belong to $LC_q$, respectively.

Fuzzy dependency (FD) \cite{JensenR2009,HuangZ2022} is a measure of uncertainty based on belongingness, which reflects how effectively $(U,\mathscr{R}_{F'})$ characterizes $U/IND(L)$. Given $<U,F,L>$, $LC_q$ is a label class. For any $F' \subseteq F$, the FD of $L$ on $F'$ is defined as:
\begin{equation}
FD(F',L) = \frac{\sum_{x_i \in U}\Big (\bigvee^{p}_{q=1}\underline{\mathscr{R}_{F'}}(LC_q)(x_{i})\Big )}{| U |},
\label{eq:7}
\end{equation}
where $|\cdot|$ denotes the cardinality of a set.

\subsection{Fuzzy Information Entropy-based Uncertainty Measures}
\label{sec:3.2}

From the perspective of disorder in pattern classification, several fuzzy information entropy-based uncertainty measures have been developed, including fuzzy entropy (FE) \cite{ZhangX2016,ZhangX2020}, fuzzy joint entropy (FJE) \cite{DaiJ2024}, fuzzy condition entropy (FCE) \cite{YangY2024}, and fuzzy mutual information (FMI) \cite{ZhangX2016,ZhangX2020}.

A $<U,F,L>$ can be transformed into $<U,F,FL>$ using Eqs.~\eqref{eq:3} and \eqref{eq:4}, and $FL$ is its fuzzy label. For any $F' \subseteq F$, FE w.r.t. $F'$, and FJE, FCE and FMI w.r.t. $F'$ and $FL$, are defined as follows:
{\small
\begin{align}
FE(F') & = -\frac{1}{| U |}\sum_{x_i \in U} \log\frac{\big | [x_i]^{fsim}_{F'} \big |}{| U |};\\
FJE(F',FL) & = -\frac{1}{| U |}\sum_{x_i \in U} \log\frac{\big | [x_i]^{fsim}_{F'}\cap[x_i]^{fsim}_{FL} \big |}{| U |};\\
FCE(FL|F') & = -\frac{1}{| U |}\sum_{x_i \in U} \log\frac{\big | [x_i]^{fsim}_{F'}\cap[x_i]^{fsim}_{FL} \big |}{\big |[x_i]^{fsim}_{F'} \big |};\\
FMI(F';FL) & = -\frac{1}{| U |}\sum_{x_i \in U} \log\frac{\big | [x_i]^{fsim}_{F'}\big | \cdot \big | [x_i]^{fsim}_{FL} \big |}{| U | \cdot \big | [x_i]^{fsim}_{F'} \cap [x_i]^{fsim}_{FL} \big |},
\label{eq:11}
\end{align}}
where $[x_i]^{fsim}_{F'}$ and $[x_i]^{fsim}_{FL}$ are the fuzzy similarity classes of $x_i \in U$ associated with $\mathscr{R}_{F'}$ and $\mathscr{R}_{FL}$, respectively.

\subsection{Fuzzy Rough Feature Selection Framework}
\label{sec:3.3}
So far, various FRFS frameworks have been proposed to address different practical needs. To analyze their commonalities and provide a more comprehensive understanding, Yang et al. \cite{YangX2018} introduced a generalized form of the qualified feature subset for a pattern classification task.

Given $<U,F,L>$ and a constraint $\rho$, a feature subset $F' \subseteq F$ is considered a $\rho$-subset if it satisfies the following conditions:
\begin{enumerate}
   \item $F'$ meets the constraint $\rho$;
   \item Any strict subset $F'' \subset F'$ does not satisfy the constraint $\rho$.
\end{enumerate}

The first condition ensures that the qualified feature subset adheres to the constraint $\rho$. The second condition guarantees that no further reduction of features from the qualified subset can maintain compliance with $\rho$. In FRFS frameworks, defining an appropriate constraint $\rho$ is crucial. Jia et al. \cite{JiaX2016} have summarized various existing constraints. Typically, the uncertainty measures discussed in Subsections \ref{sec:3.1} and \ref{sec:3.2} are effective for constructing such constraints, offering semantic interpretations from different perspectives.

Finding all possible $\rho$-subsets is an NP-hard problem, so most researchers employ heuristic search strategies to obtain an approximate solution. Yang et al. \cite{YangX2018} categorized these strategies into three types: ``addition structure'', ``deletion structure'', and ``addition-deletion structure''. Generally, the ``deletion structure'' is less effective when placing more emphasis on reducing the size of the feature set, while the ``addition-deletion structure'' is regarded as an improved version of the ``addition structure'' to better satisfy the second condition. Despite this, completely avoiding the NP-hard problem remains challenging. For simplicity, Algorithm \ref{alg:algorithm1} presents the FRFS framework using a heuristic search strategy with the ``addition structure''.

\begin{algorithm}[tb]
\caption{FRFS Framework}
\label{alg:algorithm1}
\textbf{Input}: $<U,F,L>$, constraint $\rho$, fitness function $\psi$;\\
\textbf{Output}: One $\rho$-subset $F'$;
\begin{algorithmic}[1] 
\STATE $F'\leftarrow\emptyset$;
\STATE \textbf{do}
\STATE \quad For each $f_t \in F {\setminus} F'$, compute fitness value $\psi(f_t)$;
\STATE \quad Select a $f'_t$ by considering all $\{\psi(f_t): \forall f_t \in F {\setminus} F'\}$;
\STATE \quad $F' \leftarrow F' \cup \{f'_t\}$;
\STATE \textbf{until} \textit{$F'$ meets the constraint $\rho$}
\STATE \textbf{return} $F'$.
\end{algorithmic}
\end{algorithm}

\begin{figure}[!t]
\centering
\includegraphics[width=0.35\textwidth]{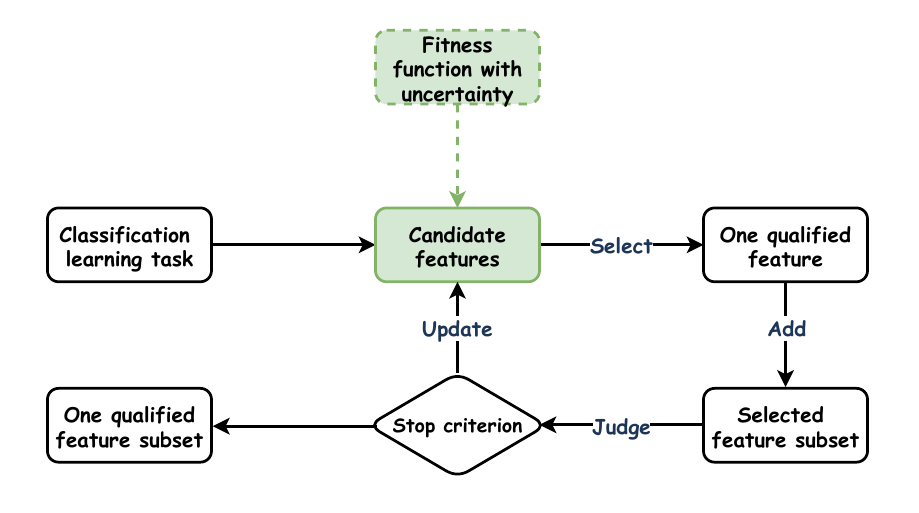}
\caption{Overall FRFS framework and underlying processing}
\label{fig:3}
\end{figure}

In Algorithm \ref{alg:algorithm1}, the choice of $f'_t$ in Step 4 depends on the fitness function $\psi$ related to the specified $\rho$. If $\rho$ is FD-related, then $\psi$ is designed as: $\psi(f_t) = FD(F' \cup \{f_t\},L) - FD(F',L)$. Here, we select a feature $f'_t$ that maximizes the increase in the FD of $L$ on $F'$ as: $f'_t = \arg \max \{\psi(f_t): \forall f_t \in F {\setminus} F'\}$. If $\rho$ consists of fuzzy information entropy-based uncertainty measures like FE, FJE, or FCE, the corresponding $\psi$ respectively are: $\psi(f_t) = FE(F' \cup \{f_t\}) - FE(F')$, $\psi(f_t) = FJE(F' \cup \{f_t\},L) - FJE(F',L)$, and $\psi(f_t) = FCE(L|F' \cup \{f_t\}) - FCE(L|F')$. In these cases, we select a feature $f'_t$ that minimally increases or maximally decreases the uncertainty as: $f'_t = \arg \min \{\psi(f_t): \forall f_t \in F {\setminus} F'\}$. It should be noted that if $\rho$ is FMI-related, the feature selection aims to maximize the increase in FMI as: $f'_t = \arg \max \{FMI(F' \cup \{f_t\};L) - FMI(F';L): \forall f_t \in F {\setminus} F'\}$. Algorithm \ref{alg:algorithm1} terminates once the selected feature subset $F'$ meets the specified $\rho$. The time complexity of Algorithm \ref{alg:algorithm1} is $\mathcal{O}(g \times |F|^2)$, where $g$ is the complexity of calculating the fitness function $\psi$. Moreover, Fig. \ref{fig:3} presents the overall FRFS framework and underlying processing.

\section{Margin-aware Fuzzy Rough Feature Selection}
\label{sec:4}

Given $<U,F,L>$, each $x_i \in U$ is an $m$-dimensional feature vector. By selecting a feature subset $F' \subseteq F$ with $m'$ features ($m' \leq m$), $x_i$ can be represented as a reduced $m'$-dimensional vector $\phi_{_{F'}}(x_i) = [f_{t_1}(x_{i}),f_{t_2}(x_{i}),\dots,f_{t_{m'}}(x_{i})]$, where $\{t_1,t_2,\ldots,t_{m'}\}$ are the indices of the selected features.

\subsection{Within-class Margin}
\label{sec:4.1}
The within-class margin measures the compactness among samples within any label class on a feature subset $F' \subseteq F$.

For a label class $LC_q \in U/IND(L)=\{LC_1, LC_2, \ldots,$ $LC_p\}$, its center on $F'$ is defined as the mean feature vector of all samples in $LC_q$:
\begin{equation}
\mu^{F'}_{q} = \frac{\sum^{|U|}_{i=1}\phi_{_{F'}}(x_i)\omega_i}{\sum^{|U|}_{i=1}\omega_i}
\label{eq:12}
\end{equation}
where $\omega_i = 1$ if $x_i \in LC_q$; otherwise, $\omega_i = 0$. The within-class margin $\vartheta^{F'}_{q}$ is calculated by the scatter of $LC_q$ as:
\begin{equation}
\vartheta^{F'}_{q} = \sum^{|U|}_{i=1}{\Vert \phi_{_{F'}}(x_i) - \mu^{F'}_{q}\Vert}_{2}\omega_i
\label{eq:13}
\end{equation}

To address label class imbalance, we normalize it as follows:
\begin{equation}
\widetilde{\vartheta^{F'}_{q}} = \frac{\sum^{|U|}_{i=1}{\Vert\phi_{_{F'}}(x_i) - \mu^{F'}_{q}\Vert}_{2}\omega_i}{\sum^{|U|}_{i=1}\omega_i}
\label{eq:14}
\end{equation}

Furthermore, for any feature subset $F' \subseteq F$, the overall within-class margin of $U/IND(L)$ on $F'$ is defined as:
\begin{equation}
\Theta^{F'}_{L} = \sum^{p}_{q=1} \widetilde{\vartheta^{F'}_{q}} = \sum^{p}_{q=1}\frac{\sum^{|U|}_{i=1}{\Vert\phi_{_{F'}}(x_i) - \mu^{F'}_{q}\Vert}_{2}\omega_i}{\sum^{|U|}_{i=1}\omega_i}
\label{eq:15}
\end{equation}

\begin{figure}[!t]
\centering
\includegraphics[width=0.24\textwidth]{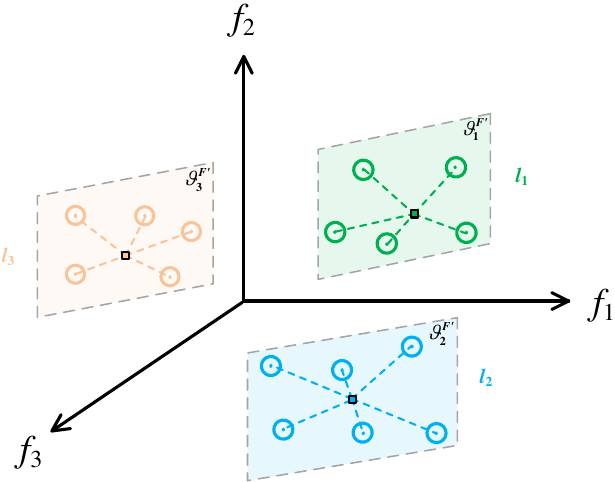}
\caption{An illustration of within-class margin}
\label{fig:4}
\end{figure}

Fig. \ref{fig:4} presents the within-class margin. As shown, the overall within-class margin $\Theta^{F'}_{L}$ on $F'$ summarizes the normalized scatter of each label class $LC_q \in U/IND(L)$ for $q = 1,2,\ldots,p$.

\subsection{Between-class Margin}
\label{sec:4.2}
The between-class margin evaluates the deviation of samples within one label class from those within other label classes on a feature subset $F' \subseteq F$. We provide two strategies to measure this margin from both global and local perspectives. 

\textbf{Global strategy:} For all label classes $U/IND(L)=\{LC_1, LC_2, \ldots,$ $LC_p\}$, their overall center on $F'$ is the mean feature vector of all samples in $U$:
\begin{equation}
\mu^{F'}_{L} = \frac{\sum^{|U|}_{i=1}\phi_{_{F'}}(x_i)}{|U|}
\label{eq:16}
\end{equation}

From a global perspective, for a label class $LC_q \in U/IND(L)$, its between-class margin $\xi^{F'}_{q}$ with respect to $U/IND(L)$ is calculated by the deviation of its center $\mu^{F'}_{q}$ from $\mu^{F'}_{L}$ as:
\begin{equation}
\xi^{F'}_{q} = {\Vert \mu^{F'}_{q} - \mu^{F'}_{L}\Vert}_{2}
\label{eq:17}
\end{equation}

Furthermore, for any feature subset $F' \subseteq F$, the global between-class margin of $U/IND(L)$ on $F'$ is defined as:
\begin{equation}
\Lambda^{F'}_{L} = \sum^{p}_{q=1}\xi^{F'}_{q} = \sum^{p}_{q=1}{\Vert \mu^{F'}_{q} - \mu^{F'}_{L}\Vert}_{2}
\label{eq:18}
\end{equation}

\begin{figure}[!t]
\centering
\subfloat[Global strategy]{\includegraphics[width=0.24\textwidth]{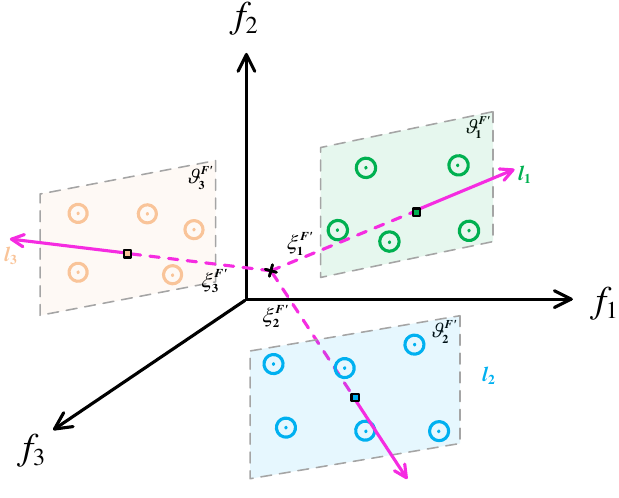}%
\label{fig:5a}}
\hfil
\subfloat[Local strategy]{\includegraphics[width=0.24\textwidth]{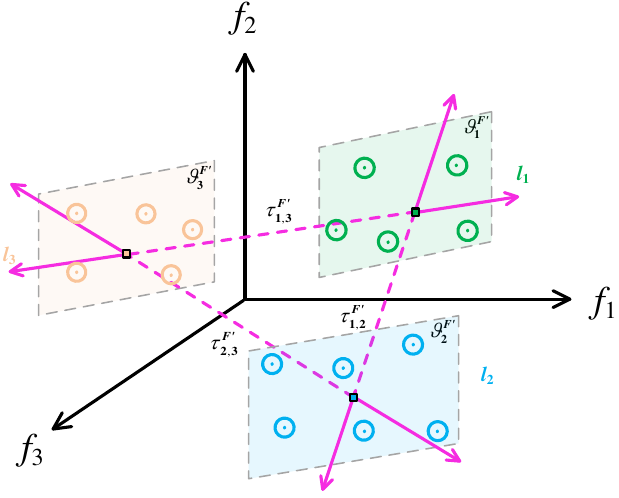}%
\label{fig:5b}}
\caption{An illustration of between-class margin}
\label{fig:5}
\end{figure}

Fig. \ref{fig:5a} presents the global between-class margin. It shows that the global between-class margin $\Lambda^{F'}_{L}$ on $F'$ summarizes the deviations of each label class center $\mu^{F'}_{q}$ from the overall center $\mu^{F'}_{L}$ for all $LC_q \in U/IND(L)$, $q = 1,2,\ldots,p$.

\textbf{Local strategy:} From a local perspective, the between-class margin $\tau^{F'}_{q,q'}$ between any two label classes $LC_q, LC_{q'} \in U/IND(L)$ is calculated by the deviation between their centers $\mu^{F'}_{q}$ and $\mu^{F'}_{q'}$ as:
\begin{equation}
\tau^{F'}_{q,q'} = {\Vert \mu^{F'}_{q} - \mu^{F'}_{q'}\Vert}_{2}
\label{eq:19}
\end{equation}

Furthermore, for any feature subset $F' \subseteq F$, the local between-class margin of $U/IND(L)$ on $F'$ is defined as:
\begin{equation}
\Delta^{F'}_{L} = \sum^{p}_{q=1}\sum^{p}_{q'=q+1}\tau^{F'}_{q,q'} = \sum^{p}_{q=1}\sum^{p}_{q'=q+1}{\Vert \mu^{F'}_{q} - \mu^{F'}_{q'}\Vert}_{2}
\label{eq:20}
\end{equation}

Fig. \ref{fig:5b} presents the local between-class margin. It shows that the local between-class margin $\Delta^{F'}_{L}$ on $F'$ summarizes the deviations between the centers of any two label classes $LC_q, LC_{q'} \in U/IND(L)$ ($q,q' = 1,2,\ldots,p$, $q' > q$).

\subsection{Within-class Between-class Margin Ratio}
\label{sec:4.3}
In a pattern classification task $<U,F,L>$, a feature subset $F' \subseteq F$ is considered beneficial for classification learning if it minimizes the within-class margin while maximizing the between-class margin. For any classifier, the difficulty of pattern classification is often closely related to both within-class and between-class margins simultaneously, and clear boundaries among different label classes are desirable. Thus, the Within-class Between-class Margin Ratio (WBMR) is proposed to integrate within-class and between-class margins.

Given a pattern classification task $<U,F,L>$ with $U/IND(L)=\{LC_1,LC_2,\ldots,LC_p\}$, $\forall F' \subseteq F$, the WBMR on $F'$ is defined as:
\begin{align}
&WBMR_{G}(F',L) = \frac{\Theta^{F'}_{L}}{\Lambda^{F'}_{L}} \nonumber\\
&= \frac{\sum^{p}_{q=1}\bigg(\sum^{|U|}_{i=1}\Big({\Vert\phi_{_{F'}}(x_i) - \mu^{F'}_{q}\Vert}_{2}\omega_i\Big)/\sum^{|U|}_{i=1}\omega_i\bigg)}{\sum^{p}_{q=1}{\Vert \mu^{F'}_{q} - \mu^{F'}_{L}\Vert}_{2}}
\label{eq:21}
\end{align}
and
\begin{align}
&WBMR_{L}(F',L) = \frac{\Theta^{F'}_{L}}{\Delta^{F'}_{L}} \nonumber\\
&= \frac{\sum^{p}_{q=1}\bigg(\sum^{|U|}_{i=1}\Big({\Vert\phi_{_{F'}}(x_i) - \mu^{F'}_{q}\Vert}_{2}\omega_i\Big)/\sum^{|U|}_{i=1}\omega_i\bigg)}{\sum^{p}_{q=1}\sum^{p}_{q'=q+1}{\Vert \mu^{F'}_{q} - \mu^{F'}_{q'}\Vert}_{2}}
\label{eq:22}
\end{align}
where $\phi_{_{F'}}(x_i)$ is an $m'$-dimensional feature vector of $x_i \in U$ in terms of the feature subset $F'$. $\mu^{F'}_{q}$, $\mu^{F'}_{q'}$ and $\mu^{F'}_{L}$ are the centers of label classes $LC_q$, $LC_{q'}$, and the overall $U/IND(L)$ on $F'$, respectively. Besides, $\omega_i = 1$ if $x_i \in LC_q$; otherwise, $\omega_i = 0$.

$WBMR_{G}(F',L)$ and $WBMR_{L}(F',L)$ represent the WBMR with global and local between-class margins, respectively. A lower WBMR suggests that the within-class margin of $U/IND(L)$ is relatively small compared to the between-class margin of $U/IND(L)$. This implies clearer class boundaries, potentially reducing the classification difficulty for various classifiers.

\subsection{Margin-aware Fuzzy Rough Feature Selection Framework}
\label{sec:4.4}
To enhance the performance of classifiers in FRFS framework, we consider the variations in both within-class and between-class margins. Using WBMR, we propose a criterion to quantitatively evaluate these margin changes, aiming to identify feature subsets $F' \subseteq F$ that can effectively balance compactness within classes and separation between classes.

Given a pattern classification task $<U,F,L>$ with $U/IND(L)=\{LC_1,LC_2,\ldots,LC_p\}$, $\forall F' \subseteq F$, if a candidate feature $f_t \in F {\setminus} F'$ is further introduced into $F'$, the variations in both within-class and between-class margins are evaluated as follows:
\begin{equation}
\varpi(f_t) = WBMR(F' \cup \{f_t\},L) - WBMR(F',L)
\label{eq:23}
\end{equation}
where $WBMR(\cdot,\cdot)$ can be specified as either $WBMR_{G}(\cdot,\cdot)$ or $WBMR_{L}(\cdot,\cdot)$ based on the type of between-class margin considered.

Based on this, in Algorithm \ref{alg:algorithm2}, we present a Margin-aware Fuzzy Rough Feature Selection (MAFRFS) framework, which similarly adopts a heuristic search strategy following the ``addition structure''.

\begin{algorithm}[tb]
\caption{MAFRFS Framework}
\label{alg:algorithm2}
\textbf{Input}: $<U,F,L>$, number of target features $k$, fitness function $\psi$, size of pool $sop$;\\
\textbf{Output}: One $\rho$-subset $F'$;
\begin{algorithmic}[1] 
\STATE $F'\leftarrow\emptyset$;
\STATE \textbf{do}
\STATE \quad $F\leftarrow F {\setminus} F'$ and $POOL\leftarrow\emptyset$;
\STATE \quad \textbf{do}
\STATE \quad For each $f_t \in F {\setminus} POOL$, compute fitness value $\psi(f_t)$;
\STATE \quad Select a $f'_t$ by considering all \\
\centerline{$\{\psi(f_t): \forall f_t \in F {\setminus} POOL\}$;}
\STATE \quad $POOL \leftarrow POOL \cup \{f'_t\}$;
\STATE \quad \textbf{until} $|POOL| = sop$
\STATE \quad For each $f_t \in POOL$, compute $\varpi(f_t)$ for evaluating\\ 
       \quad variations in within/between-class margins as Eq.~\eqref{eq:23};
\STATE \quad Select a $f'_t$ as \\
\centerline{$f'_t = \arg \min \{\varpi(f_t): \forall f_t \in POOL\}$;}
\STATE \quad $F' \leftarrow F' \cup \{f'_t\}$;
\STATE \textbf{until} $|F'| = k$
\STATE \textbf{return} $F'$.
\end{algorithmic}
\end{algorithm}

In Algorithm \ref{alg:algorithm2}, the selection of feature $f'_t$ in Step 11 is determined not only by the employed fitness function $\psi$ but also by the variations in within-class and between-class margins, as evaluated by the criterion $\varpi$. Similar to Step 3 of Algorithm \ref{alg:algorithm1}, the fitness value $\psi(f_t)$ is initially computed for each candidate feature $f_t$. However, unlike the direct selection of a single optimal feature in Step 4 of Algorithm \ref{alg:algorithm1}, Algorithm \ref{alg:algorithm2} iteratively constructs a pool of preferred candidate features $POOL$ with a size of $sop$ through Steps 5--7. For instance, if the fitness function in Step 5 is $\psi(f_t) = FD(F' \cup POOL \cup \{f_t\},L) - FD(F' \cup POOL,L)$, a feature $f'_t \in F{\setminus} \{F' \cup POOL\}$ that maximizes the increase in fuzzy dependency is included in $POOL$. Similarly, alternative fitness functions can be employed to select specific features into $POOL$ by maximizing or minimizing a given uncertainty measure. Notably, irrespective of the specific fitness function $\psi$, this process is iteratively repeated $sop$ times to construct an entire pool of candidate features.

\begin{figure}[!t]
\centering
\includegraphics[width=0.48\textwidth]{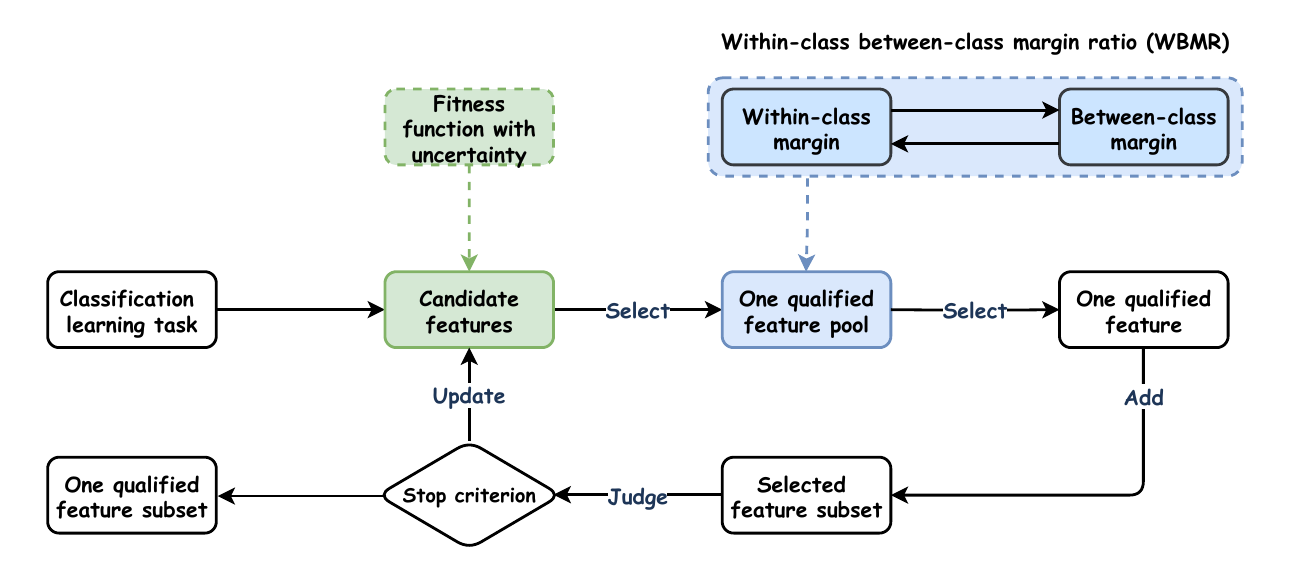}
\caption{Overall MAFRFS framework and underlying processing}
\label{fig:6}
\end{figure}

After constructing the pool of candidate features $POOL$, we further assess each element in $POOL$ for its possible impact on within-class and between-class margins of the pattern classification, as described in Step 9 of Algorithm \ref{alg:algorithm2}. Ideally, the already-selected feature subset $F'$ should possess the following desirable traits: minimal margins between samples of the same label class while maximizing the margins between samples from distinct label classes. These properties generally contribute to enhancing the performance of any classifier. Therefore, using $WBMR_{G}(F',L)$ and $WBMR_{L}(F',L)$ defined in Eqs.~\eqref{eq:21} and ~\eqref{eq:22}, we select a feature $f'_t$ from $POOL$ into the existing feature subset $F'$ based on the minimum evaluation value $\varpi(f'_t)$. Algorithm \ref{alg:algorithm2} terminates when the number of selected features reaches the predetermined limit $k$. The time complexity of Algorithm \ref{alg:algorithm2} is $\mathcal{O}(sop \times g \times |F|^2)$. In practice, when $sop = \{2,3,4\}$, the actual running time of the MAFRFS framework does not increase significantly compared to that of the FRFS framework. Fig. \ref{fig:6} presents the overall MAFRFS framework and underlying processing.

\section{Experimental Analysis}

\subsection{Datasets}
To evaluate the effectiveness of proposed MAFRFS framework, we
employ $15$ publicly available datasets from the repository\footnote{\url{https://archive.ics.uci.edu/datasets}}. Tab.~\ref{tab:1} summarizes some statistics of these datasets, with $\#Inst.$, $\#Feat.$, and $\#Lab.$ denoting the number of instances, features, and labels, respectively.

\begin{table}[!t]
\caption{Statistics of the experimental datasets.}
\centering
\setlength{\tabcolsep}{4pt}
\begin{tabular}{ccccc}
\hline
ID & Dataset & \#Inst. & \#Feat. & \#Lab. \\
\hline
 $1$ & Avila & $20$,$867$ & $10$ & $12$\\
 $2$ & Bank Marketing   & $4$,$521$ & $16$ & $2$\\
 $3$ & Cardiotocography & $2$,$126$ & $21$ & $10$\\
 $4$ & Connectionist Bench & $208$ & $60$ & $2$\\
 $5$ & Gesture Phase Segmentation & $9$,$901$ & $18$ & $5$\\
 $6$ & HCV for Egyptian patients & $1$,$385$ & $28$ & $4$\\
 $7$ & Hill-Valley & $1$,$212$ & $100$ & $2$\\
 $8$ & Leaf & $340$ & $15$ & $30$\\
 $9$ & Libras Movement & $360$ & $90$ & $15$\\
$10$ & MAGIC Gamma Telescope & $19$,$020$ & $10$ & $2$\\
$11$ & MEU-Mobile KSD & $2$,$856$ & $71$ & $56$\\
$12$ & Stalog (Landsat Satellite) & $6$,$435$ & $36$ & $6$\\
$13$ & Statlog (Vehicle Silhouettes) & $846$ & $18$ & $4$\\
$14$ & Waveform Database Generator & $5$,$000$ & $21$ & $3$\\
$15$ & Wine Quality\_red & $1$,$599$ & $11$ & $6$\\
\hline
\end{tabular}
\label{tab:1}
\end{table}

\subsection{Baselines and Settings}

We first integrate several well-established uncertainty measures into both the FRFS and MAFRFS frameworks to clearly illustrate the effectiveness and advantage of the MAFRFS framework compared to the FRFS framework. These uncertainty measures are briefly described below.

\begin{itemize}
  \item \textbf{FD} (2009) \cite{JensenR2009,HuangZ2022}: it reflects the belongingness of samples to their respective label classes. A higher FD indicates lower classification uncertainty.
  \item \textbf{FCE} (2024) \cite{YangY2024}: it characterizes the average residual uncertainty per sample within the label space when the feature space is determined. A lower FCE indicates lower uncertainty or impurity in classification.
  \item \textbf{FE} (2020) \cite{ZhangX2016,ZhangX2020}: it describes the average amount of information per sample within the feature space. A lower FE implies reduced uncertainty in classification.
  \item \textbf{FJE} (2024) \cite{DaiJ2024}: it describes the average amount of information per sample within the combination of the feature and label spaces. A lower FJE suggests reduced classification uncertainty.
  \item \textbf{FMI} (2020) \cite{ZhangX2016,ZhangX2020}: it measures how much knowing one of the feature space and the label space reduces uncertainty about the other. A higher FMI indicates lower classification uncertainty.
  \item \textbf{MFCE} (Monotonic Fuzzy Condition Entropy, 2016) \cite{ZhangX2016}: it quantifies the average residual uncertainty per sample within the label space once the feature space is established, and meets the common requirement of monotonicity. A lower MFCE indicates lower uncertainty in classification.
  \item \textbf{IPD} (Inner Product Dependency, 2022) \cite{WangC2022}: it not only considers the maximum fuzzy positive region but also minimizes classification error by accounting for the overlap among different label classes. A lower IPD signifies less uncertainty in classification.
  \item \textbf{RDSI} (Relative Decision Self-Information, 2021) \cite{WangC2021}: it simultaneously considers both the lower and upper approximations of a fuzzy label. A lower RDSI implies reduced classification uncertainty.
\end{itemize}

Next, we select two competitive combinations--each consisting of an uncertainty measure integrated into the MAFRFS framework--to demonstrate their superiority over several state-of-the-art rough-set-based feature selection algorithms, including FDM (2009) \cite{JensenR2009}, MDP (2018) \cite{DaiJ2018}, SFSS (2022) \cite{HuM2022}, N3Y (2023) \cite{LiuK2023ASOC}, FSNMER (2024) \cite{WuS2024}, and ARDSAQ (2024) \cite{QianD2024}.

It is worth noting that different algorithms may produce feature subsets of varying sizes. In general, the number of selected features impacts the performance of pattern classification: selecting too few features may lead to insufficient learnable information, whereas selecting too many may introduce redundant or irrelevant information. To ensure fairness, we modify all algorithms to output a ranking of features according to their importance for each pattern classification task.

In experimental analysis, we employ a 10-fold cross-validation, meaning that each fold consists of $90\%$ of the dataset as the training set and the remaining $10\%$ as the testing set. Feature selection algorithms are applied exclusively to the training set in each fold, producing a ranked list of features. Subsequently, we evaluate classification performance on the testing set using three classifiers: CART, SVM (-h 0 -b 1), and KNN (K = $5$). Specifically, given a dataset containing $m$ features and a ranking obtained from feature selection within each training fold, we progressively classify the testing set by gradually considering an increasing number of top-ranked features. This means performance is evaluated separately for feature combinations including: [the top $1$ ranked feature]; [the top $2$ ranked features]; [the top $3$ ranked features]; and so forth, until finally [all top $m$ ranked features]. Furthermore, the size of the pool $sop$ in MAFRFS is chosen from $\{2,3,4\}$, and the strategies employed to measure the between-class margin depend on the pattern classification tasks.

\subsection{Predictive Results}
In Tabs.~\ref{tab:2} -- \ref{tab:4}, we present comparative results of predictive performance evaluated using CART, SVM, and KNN classifiers, respectively. The reported performance values are averaged across four different proportions (30\%, 50\%, 70\%, and 90\%) of the ranked features. We integrate each of the eight uncertainty measures (FD, FCE, FE, FJE, FMI, MFCE, IPD, and RDSI) into both the MAFRFS and FRFS frameworks. The resulting comparisons are organized into eight groups, each corresponding to one uncertainty measure. Within each group, the uncertainty measure's name followed by a ``+'' symbol (e.g., FD+) indicates the predictive performance associated with feature subsets selected by the MAFRFS framework with that uncertainty measure. In contrast, the uncertainty measure's name alone (e.g., FD) denotes the predictive performance from the FRFS framework employing the same uncertainty measure. The best predictive performance between the MAFRFS and FRFS frameworks for each dataset is highlighted in boldface.

As shown in Tab.~\ref{tab:2} (CART), employing eight uncertainty measures, the MAFRFS framework (FD+, FCE+, FE+, FJE+, FMI+, MFCE+, IPD+, and RDSI+) consistently outperforms the FRFS framework across the majority of datasets. Specifically, the average performance (Ave.) demonstrates consistent improvements for each uncertainty measure within the MAFRFS framework, with growth ratios (GRs) ranging from 0.748\% (FD+ vs. FD) to 3.773\% (FE+ vs. FE). Among all results obtained by the MAFRFS framework, FE+ and FJE+ achieve relatively greater improvements over their respective baseline counterparts (FE and FJE), yielding GRs of 3.773\% and 2.984\%, respectively. Moreover, FMI+ and FCE+ also exhibit notable performance improvements exceeding 1.0\%.

In Tab.~\ref{tab:3} (SVM), the results demonstrate similar advantages for the MAFRFS framework, although several uncertainty measures integrated within the FRFS framework also yield competitive performances on certain datasets. Despite minor performance decrements observed in some comparisons (e.g., a 0.425\% decrease for FCE+ vs. FCE, and FMI+ vs. FMI), most uncertainty measures combined with the MAFRFS framework still exhibit clear average improvements, highlighting the robustness of MAFRFS. Particularly notable are FE+ and FJE+, which again achieve substantial GRs of 5.456\% and 5.060\%, respectively. These outcomes emphasize the importance and effectiveness of simultaneously accounting for variations in both within-class and between-class margins during feature selection. When employing KNN (Tab.~\ref{tab:4}), the MAFRFS framework consistently maintains superior performance. Specifically, integrating MAFRFS with all uncertainty measures yields clear average improvements relative to the FRFS. Notably, FE+ and FJE+ achieve notable GRs of 4.752\% and 3.461\%, respectively. Similarly, MFCE+ and IPD+ also provide considerable improvements (3.473\% and 3.119\%, respectively). These findings confirm that the MAFRFS framework delivers meaningful and reliable performance enhancements across diverse classifiers, highlighting its robust utility and broad applicability.

\begin{table*}[!t]
\caption{Average predictive performance (\texttt{CART}) using the top 30\%, 50\%, 70\%, and 90\% of ranked features.}
\centering
\resizebox{0.95\textwidth}{!}{
\setlength{\tabcolsep}{2pt}
\begin{tabular}{c|cc|cc|cc|cc|cc|cc|cc|cc}
\hline
\textbf{ID} & \textbf{FD+} & \textbf{FD} & \textbf{FCE+} & \textbf{FCE} & \textbf{FE+} & \textbf{FE} & \textbf{FJE+} & \textbf{FJE} & \textbf{FMI+} & \textbf{FMI} & \textbf{MFCE+} & \textbf{MFCE} & \textbf{IPD+} & \textbf{IPD} & \textbf{RDSI+} & \textbf{RDSI}\\
\hline
1 & \textbf{99.554} & 99.479 & \textbf{90.192} & 84.582 & \textbf{80.562} & 79.809 & \textbf{80.472} & 79.773 & \textbf{90.192} & 84.582 & \textbf{99.358} & 97.981 & \textbf{99.358} & 97.297 & \textbf{97.396} & 97.365\\ 
2 & 86.657 & \textbf{86.757} & \textbf{87.803} & 87.443 & \textbf{85.712} & 85.135 & \textbf{85.689} & 85.223 & \textbf{87.803} & 87.443 & \textbf{88.304} & 87.902 & \textbf{88.393} & 87.924 & 86.945 & \textbf{87.085}\\ 
3 & \textbf{73.859} & 71.254 & \textbf{70.540} & 69.728 & \textbf{55.081} & 48.574 & \textbf{60.668} & 53.884 & \textbf{70.540} & 69.728 & \textbf{66.684} & 61.081 & \textbf{70.452} & 66.216 & \textbf{68.553} & 65.828\\ 
4 & \textbf{73.880} & 72.234 & \textbf{72.284} & 71.992 & \textbf{62.843} & 62.326 & 63.594 & \textbf{63.723} & \textbf{72.284} & 71.992 & 70.503 & \textbf{71.066} & \textbf{74.603} & 71.227 & \textbf{73.897} & 71.943\\ 
5 & \textbf{75.580} & 75.469 & \textbf{83.329} & 83.186 & \textbf{72.642} & 69.820 & \textbf{71.786} & 69.928 & \textbf{83.329} & 83.186 & 90.969 & \textbf{91.413} & \textbf{82.994} & 82.930 & \textbf{75.532} & 75.320\\ 
6 & \textbf{25.020} & 24.987 & \textbf{25.443} & 24.803 & \textbf{25.396} & 24.886 & \textbf{25.341} & 25.084 & \textbf{25.443} & 24.803 & \textbf{25.096} & 24.928 & \textbf{25.112} & 25.002 & 25.171 & \textbf{25.403}\\ 
7 & \textbf{54.430} & 53.227 & \textbf{59.625} & 58.927 & 52.730 & \textbf{52.748} & 52.756 & \textbf{52.766} & \textbf{59.625} & 58.927 & \textbf{57.768} & 57.170 & \textbf{56.297} & 55.985 & \textbf{54.072} & 53.244\\ 
8 & 39.000 & \textbf{39.253} & 43.565 & \textbf{43.648} & \textbf{45.858} & 43.447 & 45.092 & \textbf{45.219} & 43.565 & \textbf{43.648} & \textbf{37.899} & 36.501 & \textbf{39.348} & 38.699 & 38.919 & \textbf{39.181}\\ 
9 & \textbf{60.600} & 58.104 & 50.118 & \textbf{50.623} & \textbf{47.247} & 45.601 & \textbf{48.884} & 48.654 & 50.118 & \textbf{50.623} & 53.625 & \textbf{53.696} & \textbf{58.285} & 56.728 & \textbf{60.543} & 58.159\\ 
10& 76.104 & \textbf{76.123} & \textbf{76.136} & 75.965 & \textbf{70.134} & 67.862 & \textbf{71.799} & 68.710 & \textbf{76.136} & 75.965 & \textbf{75.769} & 73.252 & \textbf{75.769} & 73.176 & \textbf{76.077} & 75.210\\ 
11& 35.848 & \textbf{38.108} & \textbf{57.287} & 55.781 & \textbf{40.227} & 39.512 & \textbf{45.393} & 43.688 & \textbf{57.287} & 55.781 & \textbf{28.572} & 26.770 & \textbf{44.980} & 44.155 & \textbf{39.006} & 36.772\\ 
12& \textbf{82.479} & 82.290 & \textbf{82.268} & 82.162 & \textbf{68.044} & 65.282 & \textbf{75.673} & 71.033 & \textbf{82.268} & 82.162 & \textbf{81.417} & 80.652 & 82.470 & \textbf{82.513} & \textbf{82.574} & 80.683\\ 
13& 57.577 & \textbf{57.871} & \textbf{63.511} & 63.242 & \textbf{62.486} & 59.602 & \textbf{62.439} & 59.440 & \textbf{63.511} & 63.242 & \textbf{57.180} & 54.706 & \textbf{59.088} & 58.491 & \textbf{57.495} & 54.387\\ 
14& \textbf{65.765} & 64.999 & \textbf{69.836} & 69.752 & \textbf{58.603} & 52.653 & \textbf{63.701} & 61.403 & \textbf{69.836} & 69.752 & \textbf{68.851} & 68.446 & 67.763 & \textbf{67.842} & \textbf{69.661} & 69.088\\ 
15& \textbf{55.225} & 54.285 & \textbf{57.327} & 57.291 & \textbf{53.648} & 51.919 & \textbf{53.641} & 52.119 & \textbf{57.327} & 57.291 & \textbf{55.373} & 54.228 & \textbf{57.036} & 56.469 & \textbf{57.023} & 56.057\\
\hline
Ave. & \textbf{64.105} & 63.629 & \textbf{65.951} & 65.275 & \textbf{58.747} & 56.611 & \textbf{60.462} & 58.710 & \textbf{65.951} & 65.275 & \textbf{63.824} & 62.653 & \textbf{65.463} & 64.310 & \textbf{64.191} & 63.048\\
\hline
GR & $\uparrow$ & 0.748\% & $\uparrow$ & 1.036\% & $\uparrow$ & 3.773\% & $\uparrow$ & 2.984\% & $\uparrow$ & 1.036\% & $\uparrow$ & 1.870\% & $\uparrow$ & 1.792\% & $\uparrow$ & 1.812\% \\
\hline
\end{tabular}
}
\label{tab:2}
\end{table*}

\begin{table*}[!t]
\caption{Average predictive performance (\texttt{SVM}) using the top 30\%, 50\%, 70\%, and 90\% of ranked features.}
\centering
\resizebox{0.95\textwidth}{!}{
\setlength{\tabcolsep}{2pt}
\begin{tabular}{c|cc|cc|cc|cc|cc|cc|cc|cc}
\hline
\textbf{ID} & \textbf{FD+} & \textbf{FD} & \textbf{FCE+} & \textbf{FCE} & \textbf{FE+} & \textbf{FE} & \textbf{FJE+} & \textbf{FJE} & \textbf{FMI+} & \textbf{FMI} & \textbf{MFCE+} & \textbf{MFCE} & \textbf{IPD+} & \textbf{IPD} & \textbf{RDSI+} & \textbf{RDSI}\\
\hline
1 & \textbf{52.587} & 52.422 & 53.385 & \textbf{54.179} & \textbf{49.488} & 44.294 & \textbf{49.570} & 44.325 & 53.385 & \textbf{54.179} & \textbf{52.881} & 50.872 & \textbf{52.736} & 50.838 & \textbf{51.191} & 49.531\\
2 & \textbf{88.931} & 88.755 & \textbf{89.227} & 89.096 & \textbf{88.895} & 88.741 & \textbf{88.973} & 88.741 & \textbf{89.227} & 89.096 & \textbf{89.011} & 88.787 & \textbf{89.050} & 88.761 & \textbf{88.981} & 88.874\\
3 & \textbf{70.368} & 67.468 & \textbf{69.033} & 68.754 & \textbf{51.256} & 45.561 & \textbf{58.987} & 51.807 & \textbf{69.033} & 68.754 & \textbf{60.723} & 53.930 & \textbf{66.581} & 60.645 & \textbf{61.558} & 58.043\\
4 & \textbf{73.269} & 72.395 & \textbf{73.067} & 72.705 & \textbf{64.627} & 61.625 & \textbf{66.133} & 62.045 & \textbf{73.067} & 72.705 & \textbf{67.820} & 67.069 & \textbf{73.945} & 71.383 & \textbf{72.260} & 71.945\\
5 & \textbf{50.507} & 49.632 & 51.003 & \textbf{51.331} & \textbf{46.306} & 43.550 & \textbf{45.718} & 42.380 & 51.003 & \textbf{51.331} & \textbf{48.437} & 46.512 & \textbf{51.684} & 51.374 & \textbf{50.335} & 49.828\\
6 & 23.308 & \textbf{24.143} & \textbf{23.782} & 23.569 & \textbf{25.806} & 25.468 & \textbf{25.682} & 25.431 & \textbf{23.782} & 23.569 & \textbf{24.686} & 24.197 & \textbf{23.835} & 23.442 & 23.522 & \textbf{24.947}\\
7 & \textbf{46.638} & 46.626 & 46.999 & \textbf{47.016} & \textbf{46.389} & 46.234 & \textbf{46.392} & 46.235 & 46.999 & \textbf{47.016} & \textbf{46.738} & 46.709 & \textbf{46.791} & 46.721 & \textbf{46.645} & 46.623\\
8 & \textbf{20.154} & 19.230 & 25.728 & \textbf{26.206} & \textbf{22.665} & 19.960 & \textbf{24.728} & 23.644 & 25.728 & \textbf{26.206} & \textbf{19.661} & 17.798 & \textbf{19.639} & 18.628 & \textbf{19.901} & 18.875\\
9 & \textbf{48.098} & 46.902 & 36.416 & \textbf{36.805} & \textbf{31.708} & 30.387 & \textbf{34.943} & 34.068 & 36.416 & \textbf{36.805} & \textbf{38.189} & 37.721 & \textbf{44.882} & 42.546 & \textbf{49.166} & 48.353\\
10& 80.081 & \textbf{80.118} & \textbf{78.580} & 78.469 & \textbf{73.519} & 72.098 & \textbf{73.995} & 72.367 & \textbf{78.580} & 78.469 & \textbf{79.426} & 78.210 & \textbf{79.426} & 78.212 & \textbf{80.084} & 79.486\\
11& \textbf{28.791} & 27.880 & 20.054 & \textbf{22.501} & \textbf{12.667} & 11.973 & \textbf{13.238} & 12.605 & 20.054 & \textbf{22.501} & \textbf{24.545} & 22.815 & \textbf{28.581} & 27.191 & \textbf{30.207} & 28.061\\
12& \textbf{82.135} & 81.902 & \textbf{82.121} & 81.979 & \textbf{66.440} & 62.453 & \textbf{73.700} & 67.974 & \textbf{82.121} & 81.979 & \textbf{81.767} & 81.026 & \textbf{82.624} & 82.541 & \textbf{82.479} & 80.708\\
13& 49.136 & \textbf{50.100} & 54.969 & \textbf{55.620} & \textbf{50.303} & 48.766 & \textbf{50.680} & 48.962 & 54.969 & \textbf{55.620} & \textbf{49.363} & 47.793 & \textbf{50.951} & 50.442 & \textbf{49.463} & 48.111\\
14& \textbf{73.409} & 72.624 & \textbf{77.331} & 76.999 & \textbf{67.625} & 61.229 & \textbf{71.514} & 69.498 & \textbf{77.331} & 76.999 & \textbf{76.551} & 76.325 & \textbf{75.592} & 75.345 & \textbf{77.008} & 76.286\\
15& \textbf{54.079} & 53.396 & 56.375 & \textbf{56.418} & \textbf{51.125} & 47.742 & \textbf{51.357} & 48.169 & 56.375 & \textbf{56.418} & \textbf{53.916} & 53.388 & \textbf{56.579} & 55.926 & \textbf{56.029} & 55.170\\
\hline
Ave. & \textbf{56.099} & 55.573 & 55.871 & \textbf{56.110} & \textbf{49.921} & 47.339 & \textbf{51.707} & 49.217 & 55.871 & \textbf{56.110} & \textbf{54.248} & 52.877 & \textbf{56.193} & 54.933 & \textbf{55.922} & 54.989 \\
\hline
GR & $\uparrow$ & 0.947\% & $\downarrow$ & -0.425\% & $\uparrow$ & 5.456\% & $\uparrow$ & 5.060\% & $\downarrow$ & -0.425\% & $\uparrow$ & 2.592\% & $\uparrow$ & 2.294\% & $\uparrow$ & 1.696\% \\ 
\hline
\end{tabular}
}
\label{tab:3}
\end{table*}

\begin{table*}[!t]
\caption{Average predictive performance (\texttt{KNN}) using the top 30\%, 50\%, 70\%, and 90\% of ranked features.}
\centering
\resizebox{0.95\textwidth}{!}{
\setlength{\tabcolsep}{2pt}
\begin{tabular}{c|cc|cc|cc|cc|cc|cc|cc|cc}
\hline
\textbf{ID} & \textbf{FD+} & \textbf{FD} & \textbf{FCE+} & \textbf{FCE} & \textbf{FE+} & \textbf{FE} & \textbf{FJE+} & \textbf{FJE} & \textbf{FMI+} & \textbf{FMI} & \textbf{MFCE+} & \textbf{MFCE} & \textbf{IPD+} & \textbf{IPD} & \textbf{RDSI+} & \textbf{RDSI}\\
\hline
1 & \textbf{98.275} & 94.611 & \textbf{89.181} & 82.499 & \textbf{61.027} & 58.503 & \textbf{60.953} & 58.434 & \textbf{89.181} & 82.499 & \textbf{97.700} & 82.784 & \textbf{97.700} & 83.062 & \textbf{87.794} & 81.190\\
2 & \textbf{88.182} & 88.162 & \textbf{88.968} & 88.732 & \textbf{88.409} & 87.888 & \textbf{88.469} & 87.939 & \textbf{88.968} & 88.732 & \textbf{88.324} & 87.002 & \textbf{88.535} & 88.251 & 88.323 & \textbf{88.391}\\
3 & \textbf{69.504} & 67.188 & \textbf{65.234} & 64.644 & \textbf{50.895} & 41.732 & \textbf{54.998} & 47.236 & \textbf{65.234} & 64.644 & \textbf{62.358} & 56.872 & \textbf{67.521} & 62.355 & \textbf{67.221} & 63.743\\
4 & 78.682 & \textbf{79.309} & \textbf{77.689} & 76.620 & \textbf{69.728} & 67.440 & \textbf{70.117} & 67.856 & \textbf{77.689} & 76.620 & \textbf{75.312} & 74.715 & \textbf{78.198} & 77.180 & 79.253 & \textbf{79.254}\\
5 & 77.805 & \textbf{77.812} & \textbf{74.376} & 74.207 & \textbf{74.958} & 73.210 & \textbf{74.035} & 73.218 & \textbf{74.376} & 74.207 & 82.855 & \textbf{83.948} & \textbf{76.084} & 76.034 & 77.712 & \textbf{77.730}\\
6 & 25.101 & \textbf{25.847} & \textbf{24.578} & 24.569 & \textbf{25.239} & 25.061 & \textbf{25.198} & 24.914 & \textbf{24.578} & 24.569 & \textbf{24.975} & 24.330 & \textbf{24.955} & 24.855 & 25.150 & \textbf{25.333}\\
7 & 52.505 & \textbf{52.627} & \textbf{57.144} & 57.122 & \textbf{54.434} & 54.338 & \textbf{54.495} & 54.400 & \textbf{57.144} & 57.122 & \textbf{54.272} & 53.890 & 54.177 & \textbf{54.356} & \textbf{52.678} & 52.582\\
8 & \textbf{31.646} & 31.162 & \textbf{48.182} & 47.906 & \textbf{47.669} & 42.511 & \textbf{47.309} & 47.111 & \textbf{48.182} & 47.906 & \textbf{30.942} & 30.432 & \textbf{31.386} & 30.440 & \textbf{31.588} & 30.320\\
9 & \textbf{71.882} & 70.186 & 52.680 & \textbf{52.762} & \textbf{48.825} & 46.954 & \textbf{50.808} & 50.037 & 52.680 & \textbf{52.762} & \textbf{65.881} & 65.236 & \textbf{71.112} & 69.351 & \textbf{72.266} & 70.400\\
10& \textbf{79.158} & 79.105 & \textbf{79.003} & 78.667 & \textbf{73.898} & 71.884 & \textbf{75.764} & 72.881 & \textbf{79.003} & 78.667 & \textbf{78.905} & 76.451 & \textbf{78.905} & 76.432 & \textbf{79.142} & 78.351\\
11& \textbf{34.534} & 34.507 & \textbf{47.706} & 46.887 & \textbf{36.456} & 35.597 & \textbf{42.873} & 41.176 & \textbf{47.706} & 46.887 & \textbf{28.810} & 26.801 & \textbf{43.464} & 42.615 & \textbf{37.933} & 34.372\\
12& \textbf{85.475} & 85.367 & \textbf{84.821} & 84.542 & \textbf{71.734} & 68.556 & \textbf{78.188} & 73.700 & \textbf{84.821} & 84.542 & \textbf{84.475} & 83.851 & \textbf{85.344} & 85.322 & \textbf{85.568} & 83.578\\
13& \textbf{56.729} & 55.812 & \textbf{63.365} & 63.086 & \textbf{62.865} & 60.685 & \textbf{62.742} & 60.094 & \textbf{63.365} & 63.086 & \textbf{56.390} & 54.002 & \textbf{59.645} & 57.987 & \textbf{57.119} & 54.279\\
14& \textbf{69.728} & 68.616 & \textbf{73.704} & 73.308 & \textbf{62.611} & 55.649 & \textbf{67.062} & 65.057 & \textbf{73.704} & 73.308 & \textbf{72.742} & 72.229 & 71.899 & \textbf{72.030} & \textbf{73.331} & 72.572\\
15& \textbf{53.541} & 52.566 & 56.180 & \textbf{56.317} & \textbf{52.250} & 51.025 & \textbf{52.392} & 51.060 & 56.180 & \textbf{56.317} & \textbf{53.567} & 52.824 & \textbf{55.557} & 54.437 & \textbf{55.155} & 54.704\\
\hline
Ave. & \textbf{64.850} & 64.192 & \textbf{65.521} & 64.791 & \textbf{58.733} & 56.069 & \textbf{60.360} & 58.341 & \textbf{65.521} & 64.791 & \textbf{63.834} & 61.691 & \textbf{65.632} & 63.647 & \textbf{64.682} & 63.120\\
\hline
GR & $\uparrow$ & 1.025\% & $\uparrow$ & 1.126\% & $\uparrow$ & 4.752\% & $\uparrow$ & 3.461\% & $\uparrow$ & 1.126\% & $\uparrow$ & 3.473\% & $\uparrow$ & 3.119 & $\uparrow$ & 2.475\\ 
\hline
\end{tabular}
}
\label{tab:4}
\end{table*}

\begin{figure*}[!t]
\centering
\includegraphics[width=0.9\textwidth]{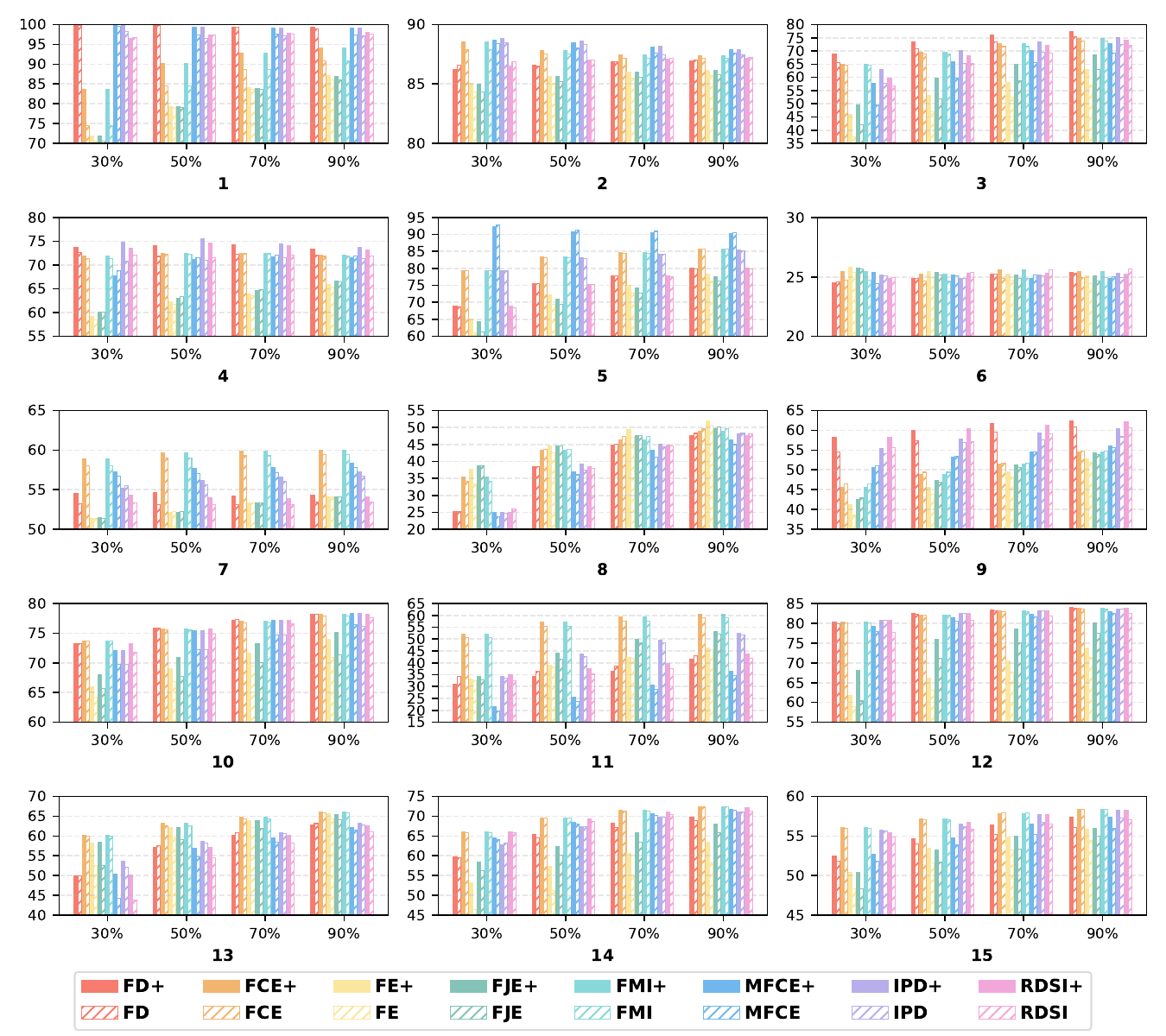}
\caption{Predictive performance (\texttt{CART}) using the top 30\%, 50\%, 70\%, and 90\% of ranked features across $15$ datasets.}
\label{fig:7}
\end{figure*}

%

In addition, Fig. \ref{fig:7} presents detailed comparisons of predictive performance when employing the top 30\%, 50\%, 70\%, and 90\% of ranked features across $15$ datasets. The predictive performance in Fig. \ref{fig:7} is evaluated using the CART classifier. Due to space limitations, comparisons using the other two classifiers (i.e., SVM and KNN) are provided in the supplementary materials. From Fig. \ref{fig:7}, we can clearly observe that integrating each of the eight uncertainty measures, i.e., FD, FCE, FE, FJE, FMI, MFCE, IPD, and RDSI, into the MAFRFS framework leads to significant improvements compared to the FRFS framework in most cases. The observation regarding the predictive performance evaluated using SVM and KNN is similar to that shown in Fig. \ref{fig:7}.

We further evaluate the superiority of the MAFRFS framework by comparing it with six state-of-the-art rough-set-based feature selection algorithms (FDM, MDP, SFSS, N3Y, FSNMER, and ARDSAQ). Note that the MAFRFS framework itself cannot independently perform feature selection without incorporating an uncertainty measure, and its performance directly depends on the effectiveness of the selected measure. In other words, \textit{MAFRFS is expected to stand on the shoulders of giants}.

\begin{table*}[!t]
\caption{Average predictive performance using the top 30\%, 50\%, 70\%, and 90\% of ranked features for eight algorithms across $15$ datasets, evaluated by \texttt{CART}, \texttt{SVM}, and \texttt{KNN}.}
\centering
\resizebox{0.95\textwidth}{!}{
\setlength{\tabcolsep}{2pt}
\begin{tabular}{ccccccccccccccccc}
\hline
\textbf{Cls.} & \textbf{SOTA} & \textbf{1} & \textbf{2} & \textbf{3} & \textbf{4} & \textbf{5} & \textbf{6} & \textbf{7} & \textbf{8} & \textbf{9} & \textbf{10} & \textbf{11} & \textbf{12} & \textbf{13} & \textbf{14} & \textbf{15}\\
\hline
\multirow{8}{*}{\rotatebox{90}{\textbf{CART}}} & FDM & 86.872 & 86.612 & 75.376 & 68.870 & 83.348 & 23.592 & 54.373 & 51.691 & 58.472 & 77.535 & 53.965 & 84.553 & 66.637 & 70.390 & 58.005\\
& MDP & 82.028 & 87.099 & 77.211 & 70.433 & 87.274 & 24.170 & 55.920 & 53.750 & 64.167 & 76.735 & 45.378 & 84.312 & \textbf{69.976} & 70.300 & 57.755\\
& SFSS & 98.924 & 87.083 & 79.868 & 73.317 & 88.774 & 25.415 & 59.612 & 54.485 & 60.903 & 80.208 & 64.522 & 85.365 & 67.937 & 74.885 & \textbf{59.381}\\
& N3Y & 98.078 & 86.651 & 79.751 & \textbf{74.279} & 89.153 & 24.657 & 54.827 & 55.221 & 61.944 & 77.927 & 48.030 & 84.790 & 66.814 & \textbf{75.135} & 57.536\\
& FSNMER & 94.283 & 85.999 & 73.001 & 69.471 & 85.317 & 24.639 & 56.704 & 53.456 & 58.194 & 78.613 & 63.463 & 85.645 & 67.553 & 72.340 & 56.801\\
& ARDSAQ & \textbf{99.798} & 87.320 & 80.609 & 73.438 & \textbf{90.125} & 23.556 & 57.199 & \textbf{57.721} & 59.792 & 77.772 & 49.632 & \textbf{85.878} & 68.469 & 72.530 & 56.707\\
\cdashline{2-17}
& \textbf{FD+} & 99.276 & \textbf{87.431} & \textbf{81.056} & 73.558 & 85.481 & 25.740 & 55.301 & 56.838 & \textbf{65.417} & 80.191 & 47.041 & 85.785 & 68.972 & 74.305 & 58.974\\
& \textbf{FCE+} & 99.358 & 87.038 & 78.692 & 73.077 & 88.882 & \textbf{26.444} & \textbf{61.015} & 53.529 & 59.028 & \textbf{80.210} & \textbf{64.758} & 85.338 & 69.031 & 75.000 & 59.256\\
\hline
\multirow{8}{*}{\rotatebox{90}{\textbf{SVM}}} & FDM & 50.252 & 88.703 & 68.826 & 70.793 & 52.896 & 23.285 & 46.700 & 22.206 & 41.597 & 79.819 & 20.860 & 84.114 & 53.576 & 79.280 & 56.113\\
& MDP & 52.371 & 89.405 & 70.814 & 72.716 & 53.217 & 24.549 & 46.658 & 24.559 & 46.875 & 79.829 & 24.930 & 84.297 & \textbf{59.279} & 79.055 & 56.223\\
& SFSS & 57.089 & \textbf{89.516} & 74.730 & 73.197 & 53.341 & 23.484 & 46.865 & \textbf{30.662} & 45.139 & 80.578 & 19.249 & 84.930 & 54.551 & \textbf{84.455} & 57.114\\
& N3Y & 57.418 & 89.200 & 75.223 & 73.077 & 54.353 & 24.585 & 47.339 & 30.221 & 45.625 & 81.288 & \textbf{31.985} & 84.981 & 55.467 & 83.310 & 57.004\\
& FSNMER & 51.409 & 88.808 & 68.262 & 68.990 & 53.121 & 25.018 & 47.380 & 24.706 & 43.264 & 81.425 & 19.231 & 84.643 & 56.708 & 82.230 & 54.237\\
& ARDSAQ & 55.938 & 89.145 & 75.470 & 73.197 & 48.952 & \textbf{25.144} & \textbf{47.422} & 24.485 & 41.181 & 78.994 & 26.401 & 85.043 & 57.181 & 82.015 & 54.941\\
\cdashline{2-17}
& \textbf{FD+} & \textbf{57.682} & 89.350 & \textbf{76.399} & \textbf{73.918} & \textbf{54.742} & 23.610 & 46.700 & 25.956 & \textbf{49.931} & \textbf{82.520} & 30.917 & \textbf{85.159} & 57.181 & 83.425 & \textbf{57.161}\\
& \textbf{FCE+} & 56.635 & 89.466 & 74.741 & 73.918 & 53.944 & 23.809 & 46.906 & 28.235 & 44.375 & 80.578 & 22.330 & 84.930 & 56.619 & 84.335 & 57.036\\
\hline
\multirow{8}{*}{\rotatebox{90}{\textbf{KNN}}} & FDM & 66.581 & 88.686 & 72.472 & 80.529 & 85.529 & 23.664 & 53.527 & 47.721 & 75.208 & 80.028 & 43.697 & 89.386 & 65.573 & 74.290 & 56.770\\
& MDP & 64.555 & 88.393 & 72.919 & 81.971 & 85.491 & 24.567 & 53.342 & 46.544 & 74.375 & 79.508 & 42.918 & 89.561 & 68.115 & 73.990 & 56.332\\
& SFSS & 93.498 & 88.730 & \textbf{78.046} & 82.332 & 83.747 & 24.657 & 55.239 & 60.515 & 68.403 & 82.592 & \textbf{54.044} & 89.841 & 68.203 & \textbf{80.870} & 58.068\\
& N3Y & 83.932 & 88.244 & 75.611 & 82.212 & 85.113 & 26.300 & 52.867 & \textbf{62.574} & 75.972 & 81.046 & 46.008 & 89.926 & 66.401 & 79.795 & 57.489\\
& FSNMER & 65.320 & 88.056 & 71.225 & 79.928 & 85.327 & 25.144 & 54.435 & 51.618 & 71.181 & 81.073 & 45.825 & 89.814 & 66.696 & 78.240 & 54.503\\
& ARDSAQ & 93.037 & 87.995 & 76.517 & \textbf{82.933} & \textbf{87.931} & 24.838 & 54.249 & 49.559 & \textbf{76.181} & 80.247 & 41.115 & 90.179 & 67.228 & 77.815 & 54.659\\
\cdashline{2-17}
& \textbf{FD+} & 94.528 & 88.548 & 78.022 & 81.370 & 86.302 & \textbf{26.390} & 53.342 & 48.824 & 75.347 & \textbf{82.741} & 42.349 & \textbf{90.416} & 68.351 & 79.885 & 58.224\\
& \textbf{FCE+} & \textbf{95.337} & \textbf{88.841} & 76.929 & 82.452 & 83.729 & 25.542 & \textbf{57.261} & 59.485 & 65.278 & 82.592 & 51.558 & 89.736 & \textbf{69.385} & 80.845 & \textbf{58.412}\\
\hline
\end{tabular}
}
\label{tab:5}
\end{table*}

Thus, for comparative experiments, we select two competitive combinations, namely FD+ and FCE+. As clearly indicated in Tab.~\ref{tab:5}, FD+ and FCE+ consistently achieve better predictive performance than FDM, MDP, SFSS, N3Y, FSNMER, and ARDSAQ across three widely-used classifiers evaluated on $15$ datasets. Specifically, under the CART classifier, FD+ outperforms other algorithms on 20.00\% ($3$ out of $15$) of datasets, whereas FCE+ achieves top performance on 26.67\% ($4$ out of $15$). Under the SVM classifier, FD+ attains superior performance on 53.33\% ($8$ out of $15$) of the datasets--the highest proportion among all compared algorithms--although FCE+ does not yield the best results with SVM. Under the KNN classifier, FD+ and FCE+ secure top overall performance on 20.00\% ($3$ out of $15$) and 33.33\% ($5$ out of $15$) of the datasets, respectively. Overall, these results confirm that both FD+ and FCE+ frequently outperform the six state-of-the-art rough-set-based feature selection algorithms, clearly showcasing their effectiveness and robustness across various classifiers and datasets.

\subsection{Statistical Tests}

To compare predictive performance in a statistically well-founded way, we employ the \emph{Friedman test} \cite{FriedmanM1940}. Suppose we have $s$ algorithms to compare across $N$ datasets. Let $r_i^j$ denote the ranking of the $j$-th algorithm on the $i$-th dataset, and define the average rank of the $j$-th algorithm as: $R_j = \frac{1}{N}\sum_{i=1}^{N}r_i^j$. If the $j$-th algorithm is not applicable to the $i$-th dataset, its ranking $r_i^j$ is assigned the worst (highest numerical) rank.

\begin{table}[!t]
\caption{Friedman statistic and critical value.}
\centering
\setlength{\tabcolsep}{10pt}
\begin{tabular}{ccccc}
\hline
Classifiers & $F_F$ & CV ($\alpha = 0.05$) \\
\hline
\multicolumn{1}{c}{CART} &  \multicolumn{1}{c}{$8.0500$} & \multirow{3}{*}{$2.1044$}\\
SVM       &        $8.4275$         &                   \\
KNN       &        $5.7492$         &                   \\
\hline
\end{tabular}
\label{tab:8}
\end{table}

Under the null hypothesis (no differences among algorithms), the Friedman statistic can be approximated by a Fisher ($F$) distribution with $(s-1)$ and $(s-1)(N-1)$ degrees of freedom. Specifically, the Friedman statistic $F_F$ is defined as follows:
\begin{equation}
F_F = \frac{(N-1)\chi_F^2}{N(s-1)-\chi_F^2},
\label{eq:24}
\end{equation}
where
\begin{equation}
\chi_F^2 = \frac{12N}{s(s+1)}(\sum_{j=1}^{s}R_j^2-\frac{s(s+1)^2}{4}).
\label{eq:25}
\end{equation}

Tab.~\ref{tab:8} provides the Friedman statistics $F_F$ along with the corresponding critical value (CV) computed for three classifiers ($s = 8$, $N = 15$). At a significance level of $0.05$, the Friedman test rejects the null hypothesis that all algorithms achieve identical predictive performance. Therefore, the post-hoc \emph{Nemenyi test} is conducted to analyze which specific algorithms differ significantly. The predictive performances of two algorithms are considered significantly different if the difference between their average ranks exceeds the critical difference (CD):
\begin{equation}
CD = q_\alpha \sqrt{\frac{s(s+1)}{6N}}
\label{eq:26}
\end{equation}
where the critical value $q_\alpha$ is derived from the Studentized range statistic divided by $\sqrt{2}$. At the significance level $\alpha = 0.05$, we have $q_\alpha = 3.0310$, thereby yielding $CD = 2.7110$.

\begin{figure*}[!t]
\centering
\subfloat[\texttt{CART}]{\includegraphics[width=0.30\textwidth]{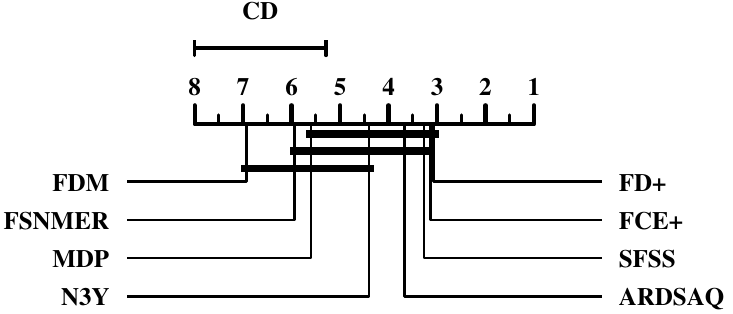}%
\label{fig:10a}}
\hfil
\subfloat[\texttt{SVM}]{\includegraphics[width=0.30\textwidth]{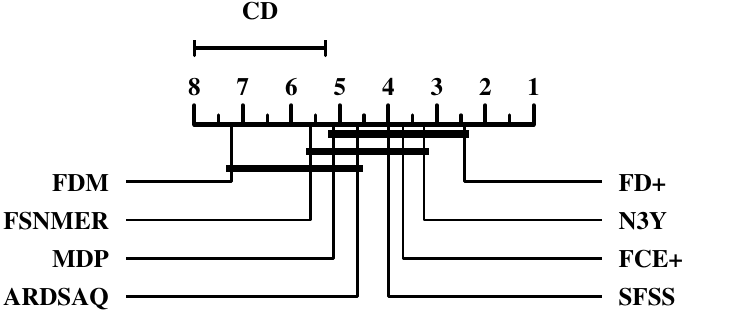}%
\label{fig:10b}}
\hfil
\subfloat[\texttt{KNN}]{\includegraphics[width=0.30\textwidth]{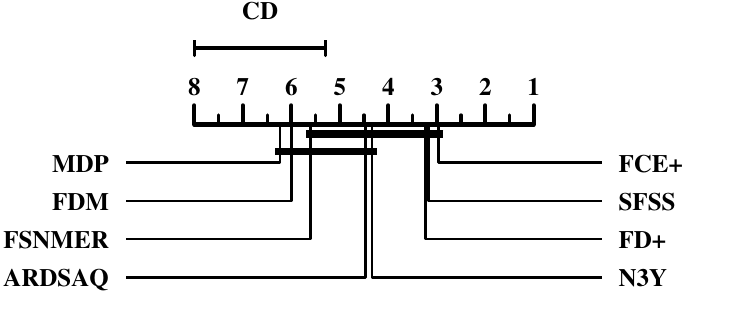}%
\label{fig:10c}}
\caption{CD diagrams for the predictive performance of eight algorithms evaluated using \texttt{CART}, \texttt{SVM}, and \texttt{KNN}.}
\label{fig:10}
\end{figure*}

To visually highlight the actual differences in predictive performance among eight algorithms, Fig~\ref{fig:10} presents CD diagrams \cite{DemvsarJ2006} that illustrate the performance of these algorithms when evaluated using three classifiers. In each subfigure, the algorithms are marked along a horizontal axis according to their average ranks, with better-performing (higher-ranked) algorithms positioned to the right. Algorithms whose predictive performances are not significantly different according to the \emph{Nemenyi test} are connected by thick horizontal lines. The corresponding CD value is indicated above each axis.

As shown in Fig.~\ref{fig:10}, out of the total $18$ predictive performance comparisons ($6$ SOTA algorithms $\times$ $3$ classifiers), our FD+ and FCE+ algorithms achieve statistically comparable performance in 72.22\% of the cases. Specifically, these include $13$ comparisons: against SFSS, ARDSAQ and N3Y across all three classifiers; against MDP under CART and SVM; and against FSNMER under SVM and KNN. Impressively, in the remaining 27.78\% of cases, FD+ and FCE+ demonstrate statistically superior predictive performance. Notably, only SFSS outperforms FD+ under the KNN classifier, while only N3Y surpasses FCE+ under the SVM classifier. These results strongly confirm the overall superior predictive capabilities of FD+ and FCE+ compared to the six state-of-the-art feature selection algorithms considered.

\section{Conclusion}
This paper proposes a two-stage feature selection framework, named MAFRFS, that simultaneously considers uncertainty reduction and enhancement of label class separability. Specifically, Within-class and between-class margins are defined, and a within-class between-class margin ratio is proposed to guide feature selection towards obtaining more separable and discriminative label class structures. Experimental results on 15 public datasets demonstrate that MAFRFS significantly outperforms FRFS and six state-of-the-art feature selection algorithms. 

\section*{Acknowledgment}

The authors would like to thank the anonymous reviewers and the editor for their constructive and valuable comments. This work is supported by the National Natural Science Foundation of China (Nos. 62076111, 51975294).

\bibliographystyle{IEEEtran}

\bibliography{bib}

\begin{thebibliography}{10}
\providecommand{\url}[1]{#1}
\csname url@samestyle\endcsname
\providecommand{\newblock}{\relax}
\providecommand{\bibinfo}[2]{#2}
\providecommand{\BIBentrySTDinterwordspacing}{\spaceskip=0pt\relax}
\providecommand{\BIBentryALTinterwordstretchfactor}{4}
\providecommand{\BIBentryALTinterwordspacing}{\spaceskip=\fontdimen2\font plus
\BIBentryALTinterwordstretchfactor\fontdimen3\font minus
  \fontdimen4\font\relax}
\providecommand{\BIBforeignlanguage}[2]{{%
\expandafter\ifx\csname l@#1\endcsname\relax
\typeout{** WARNING: IEEEtran.bst: No hyphenation pattern has been}%
\typeout{** loaded for the language `#1'. Using the pattern for}%
\typeout{** the default language instead.}%
\else
\language=\csname l@#1\endcsname
\fi
#2}}
\providecommand{\BIBdecl}{\relax}
\BIBdecl

\bibitem{WrightJ2022}
J.~Wright and Y.~Ma, \emph{High-dimensional data analysis with low-dimensional
  models: Principles, computation, and applications}.\hskip 1em plus 0.5em
  minus 0.4em\relax Cambridge University Press, 2022.

\bibitem{LaurensV2008}
L.~V.~D. Maaten and G.~Hinton, ``Visualizing data using t-sne,'' \emph{Journal
  of Machine Learning Research}, vol.~9, no.~86, pp. 2579--2605, 2008.

\bibitem{GisbrechtA2015}
A.~Gisbrecht and B.~Hammer, ``Data visualization by nonlinear dimensionality
  reduction,'' \emph{WIREs Data Mining and Knowledge Discovery}, vol.~5, no.~2,
  pp. 51--73, 2015.

\bibitem{YangQ2025}
Q.~Yang, X.~Xu, Z.~Zhan, J.~Zhong, S.~Kwong, and J.~Zhang, ``Evolutionary
  multitask optimization for multiform feature selection in classification,''
  \emph{IEEE Transactions on Cybernetics}, vol.~55, no.~4, pp. 1673--1686,
  2025.

\bibitem{ZhaoY2025}
Y.~Zhao, Z.~Bi, P.~Zhu, A.~Yuan, and X.~Li, ``Deep spectral clustering with
  projected adaptive feature selection,'' \emph{IEEE Transactions on Geoscience
  and Remote Sensing}, vol.~63, pp. 1--12, 2025.

\bibitem{ZhangX2024}
X.~Zhang, M.~Xu, and X.~Zhou, ``Realnet: A feature selection network with
  realistic synthetic anomaly for anomaly detection,'' in \emph{Proceedings of
  the IEEE/CVF Conference on Computer Vision and Pattern Recognition (CVPR)},
  June 2024, pp. 16\,699--16\,708.

\bibitem{DidierD1990}
D.~Dubois and H.~Prade, ``Rough fuzzy sets and fuzzy rough sets,''
  \emph{International Journal of General Systems}, vol.~17, no. 2-3, pp.
  191--209, 1990.

\bibitem{TsangE2008}
E.~C.~C. Tsang, D.~Chen, D.~S. Yeung, X.~Wang, and J.~W.~T. Lee, ``Attributes
  reduction using fuzzy rough sets,'' \emph{IEEE Transactions on Fuzzy
  Systems}, vol.~16, no.~5, pp. 1130--1141, 2008.

\bibitem{JensenR2009}
R.~Jensen and Q.~Shen, ``New approaches to fuzzy-rough feature selection,''
  \emph{IEEE Transactions on Fuzzy Systems}, vol.~17, no.~4, pp. 824--838,
  2009.

\bibitem{WangC2022}
C.~Wang, Y.~Qian, W.~Ding, and X.~Fan, ``Feature selection with fuzzy-rough
  minimum classification error criterion,'' \emph{IEEE Transactions on Fuzzy
  Systems}, vol.~30, no.~8, pp. 2930--2942, 2022.

\bibitem{HuangZ2022}
Z.~Huang, J.~Li, and Y.~Qian, ``Noise-tolerant fuzzy-$\beta$-covering-based
  multigranulation rough sets and feature subset selection,'' \emph{IEEE
  Transactions on Fuzzy Systems}, vol.~30, no.~7, pp. 2721--2735, 2022.

\bibitem{AnS2023}
S.~An, E.~Zhao, C.~Wang, G.~Guo, S.~Zhao, and P.~Li, ``Relative fuzzy rough
  approximations for feature selection and classification,'' \emph{IEEE
  Transactions on Cybernetics}, vol.~53, no.~4, pp. 2200--2210, 2023.

\bibitem{ZhangX2016}
X.~Zhang, C.~Mei, D.~Chen, and J.~Li, ``Feature selection in mixed data: A
  method using a novel fuzzy rough set-based information entropy,''
  \emph{Pattern Recognition}, vol.~56, pp. 1--15, 2016.

\bibitem{ZhangX2020}
X.~Zhang, C.~Mei, D.~Chen, Y.~Yang, and J.~Li, ``Active incremental feature
  selection using a fuzzy-rough-set-based information entropy,'' \emph{IEEE
  Transactions on Fuzzy Systems}, vol.~28, no.~5, pp. 901--915, 2020.

\bibitem{WanJ2021}
J.~Wan, H.~Chen, T.~Li, X.~Yang, and B.~Sang, ``Dynamic interaction feature
  selection based on fuzzy rough set,'' \emph{Information Sciences}, vol. 581,
  pp. 891--911, 2021.

\bibitem{WangZ2023}
Z.~Wang, H.~Chen, Z.~Yuan, J.~Wan, and T.~Li, ``Multiscale fuzzy entropy-based
  feature selection,'' \emph{IEEE Transactions on Fuzzy Systems}, vol.~31,
  no.~9, pp. 3248--3262, 2023.

\bibitem{WanJ2023}
J.~Wan, H.~Chen, T.~Li, Z.~Yuan, J.~Liu, and W.~Huang, ``Interactive and
  complementary feature selection via fuzzy multigranularity uncertainty
  measures,'' \emph{IEEE Transactions on Cybernetics}, vol.~53, no.~2, pp.
  1208--1221, 2023.

\bibitem{DaiJ2024}
J.~Dai, Q.~Liu, X.~Zou, and C.~Zhang, ``Feature selection based on fuzzy
  combination entropy considering global and local feature correlation,''
  \emph{Information Sciences}, vol. 652, p. 119753, 2024.

\bibitem{YangY2024}
Y.~Yang, D.~Chen, Z.~Ji, X.~Zhang, and L.~Dong, ``A two-way accelerator for
  feature selection using a monotonic fuzzy conditional entropy,'' \emph{Fuzzy
  Sets and Systems}, vol. 483, p. 108916, 2024.

\bibitem{YangJ2020}
J.~Yang, G.~Wang, Q.~Zhang, and H.~Wang, ``Knowledge distance measure for the
  multigranularity rough approximations of a fuzzy concept,'' \emph{IEEE
  Transactions on Fuzzy Systems}, vol.~28, no.~4, pp. 706--717, 2020.

\bibitem{XiaD2023}
D.~Xia, G.~Wang, Q.~Zhang, J.~Yang, H.~Bao, S.~Li, and B.~Sang, ``Interactive
  fuzzy knowledge distance-guided attribute reduction with three-way
  accelerator,'' \emph{Knowledge-Based Systems}, vol. 279, p. 110943, 2023.

\bibitem{DaiJ2018}
J.~Dai, H.~Hu, W.~Wu, Y.~Qian, and D.~Huang,
  ``Maximal-discernibility-pair-based approach to attribute reduction in fuzzy
  rough sets,'' \emph{IEEE Transactions on Fuzzy Systems}, vol.~26, no.~4, pp.
  2174--2187, 2018.

\bibitem{JiangZ2021}
Z.~Jiang, K.~Liu, J.~Song, X.~Yang, J.~Li, and Y.~Qian, ``Accelerator for
  crosswise computing reduct,'' \emph{Applied Soft Computing}, vol.~98, p.
  106740, 2021.

\bibitem{HuangW2023}
W.~Huang, Y.~She, X.~He, and W.~Ding, ``Fuzzy rough sets-based incremental
  feature selection for hierarchical classification,'' \emph{IEEE Transactions
  on Fuzzy Systems}, vol.~31, no.~10, pp. 3721--3733, 2023.

\bibitem{ZhangC2025}
C.~Zhang, Z.~Lu, Y.~Zhang, and J.~Dai, ``Online streaming feature selection
  using bidirectional complementarity based on fuzzy gini entropy,'' \emph{IEEE
  Transactions on Fuzzy Systems}, 2025, 10.1109/TFUZZ.2025.3529466.

\bibitem{XuS2016}
S.~Xu, X.~Yang, H.~Yu, D.~Yu, J.~Yang, and E.~C. Tsang, ``Multi-label learning
  with label-specific feature reduction,'' \emph{Knowledge-Based Systems}, vol.
  104, pp. 52--61, 2016.

\bibitem{WangZ2024}
Z.~Wang, D.~Chen, and X.~Che, ``Learning operator-valued kernels from
  multi-label datesets with fuzzy rough sets,'' \emph{IEEE Transactions on
  Fuzzy Systems}, 2025, 10.1109/TFUZZ.2024.3522466.

\bibitem{LiuK2023}
K.~Liu, T.~Li, X.~Yang, H.~Chen, J.~Wang, and Z.~Deng, ``Semifree:
  Semisupervised feature selection with fuzzy relevance and redundancy,''
  \emph{IEEE Transactions on Fuzzy Systems}, vol.~31, no.~10, pp. 3384--3396,
  2023.

\bibitem{ZhouN2025}
N.~Zhou, S.~Liao, H.~Chen, W.~Ding, and Y.~Lu, ``Semi-supervised feature
  selection with multi-scale fuzzy information fusion: from both global and
  local perspectives,'' \emph{IEEE Transactions on Fuzzy Systems}, 2025,
  10.1109/TFUZZ.2025.3540884.

\bibitem{WangC2019}
C.~Wang, Y.~Huang, M.~Shao, and X.~Fan, ``Fuzzy rough set-based attribute
  reduction using distance measures,'' \emph{Knowledge-Based Systems}, vol.
  164, pp. 205--212, 2019.

\bibitem{YuD2007}
D.~Yu, Q.~Hu, and C.~Wu, ``Uncertainty measures for fuzzy relations and their
  applications,'' \emph{Applied Soft Computing}, vol.~7, no.~3, pp. 1135--1143,
  2007.

\bibitem{YeJ2021}
J.~Ye, J.~Zhan, W.~Ding, and H.~Fujita, ``A novel fuzzy rough set model with
  fuzzy neighborhood operators,'' \emph{Information Sciences}, vol. 544, pp.
  266--297, 2021.

\bibitem{HuQ2011}
Q.~Hu, D.~Yu, W.~Pedrycz, and D.~Chen, ``Kernelized fuzzy rough sets and their
  applications,'' \emph{IEEE Transactions on Knowledge and Data Engineering},
  vol.~23, no.~11, pp. 1649--1667, 2011.

\bibitem{HuM2024}
M.~Hu, Y.~Guo, R.~Wang, and X.~Wang, ``Attribute reduction with fuzzy
  kernel-induced relations,'' \emph{Information Sciences}, vol. 669, p. 120589,
  2024.

\bibitem{HuQ2011FSS}
Q.~Hu, S.~An, X.~Yu, and D.~Yu, ``Robust fuzzy rough classifiers,'' \emph{Fuzzy
  Sets and Systems}, vol. 183, no.~1, pp. 26--43, 2011.

\bibitem{ZhaoS2015}
S.~Zhao, H.~Chen, C.~Li, X.~Du, and H.~Sun, ``A novel approach to building a
  robust fuzzy rough classifier,'' \emph{IEEE Transactions on Fuzzy Systems},
  vol.~23, no.~4, pp. 769--786, 2015.

\bibitem{VluymansS2016}
S.~Vluymans, D.~{Sánchez Tarragó}, Y.~Saeys, C.~Cornelis, and F.~Herrera,
  ``Fuzzy rough classifiers for class imbalanced multi-instance data,''
  \emph{Pattern Recognition}, vol.~53, pp. 36--45, 2016.

\bibitem{WangC2017}
C.~Wang, Y.~Qi, M.~Shao, Q.~Hu, D.~Chen, Y.~Qian, and Y.~Lin, ``A fitting model
  for feature selection with fuzzy rough sets,'' \emph{IEEE Transactions on
  Fuzzy Systems}, vol.~25, no.~4, pp. 741--753, 2017.

\bibitem{LiY2017}
Y.~Li, S.~Wu, Y.~Lin, and J.~Liu, ``Different classes' ratio fuzzy rough set
  based robust feature selection,'' \emph{Knowledge-Based Systems}, vol. 120,
  pp. 74--86, 2017.

\bibitem{SkowronA1992}
A.~Skowron and C.~Rauszer, \emph{The Discernibility Matrices and Functions in
  Information Systems}.\hskip 1em plus 0.5em minus 0.4em\relax Springer
  Netherlands, 1992, pp. 331--362.

\bibitem{YangX2013}
X.~Yang, Y.~Qi, X.~Song, and J.~Yang, ``Test cost sensitive multigranulation
  rough set: Model and minimal cost selection,'' \emph{Information Sciences},
  vol. 250, pp. 184--199, 2013.

\bibitem{SunL2023}
L.~Sun, S.~Si, W.~Ding, X.~Wang, and J.~Xu, ``Tfsfb: Two-stage feature
  selection via fusing fuzzy multi-neighborhood rough set with binary whale
  optimization for imbalanced data,'' \emph{Information Fusion}, vol.~95, pp.
  91--108, 2023.

\bibitem{LuoC2023}
C.~Luo, S.~Wang, T.~Li, H.~Chen, J.~Lv, and Z.~Yi, ``Large-scale meta-heuristic
  feature selection based on bpso assisted rough hypercuboid approach,''
  \emph{IEEE Transactions on Neural Networks and Learning Systems}, vol.~34,
  no.~12, pp. 10\,889--10\,903, 2023.

\bibitem{YangX2018}
X.~Yang and Y.~Yao, ``Ensemble selector for attribute reduction,''
  \emph{Applied Soft Computing}, vol.~70, pp. 1--11, 2018.

\bibitem{JiaX2016}
X.~Jia, L.~Shang, B.~Zhou, and Y.~Yao, ``Generalized attribute reduct in rough
  set theory,'' \emph{Knowledge-Based Systems}, vol.~91, pp. 204--218, 2016.

\bibitem{WangC2021}
C.~Wang, Y.~Huang, W.~Ding, and Z.~Cao, ``Attribute reduction with fuzzy rough
  self-information measures,'' \emph{Information Sciences}, vol. 549, pp.
  68--86, 2021.

\bibitem{HuM2022}
M.~Hu, E.~C.~C. Tsang, Y.~Guo, and W.~Xu, ``Fast and robust attribute reduction
  based on the separability in fuzzy decision systems,'' \emph{IEEE
  Transactions on Cybernetics}, vol.~52, no.~6, pp. 5559--5572, 2022.

\bibitem{LiuK2023ASOC}
K.~Liu, T.~Li, X.~Yang, H.~Ju, X.~Yang, and D.~Liu, ``Feature selection in
  threes: Neighborhood relevancy, redundancy, and granularity interactivity,''
  \emph{Applied Soft Computing}, vol. 146, p. 110679, 2023.

\bibitem{WuS2024}
S.~Wu, L.~Wang, S.~Ge, Z.~Hao, and Y.~Liu, ``Neighborhood rough set with
  neighborhood equivalence relation for feature selection,'' \emph{Knowledge
  and Information Systems}, vol.~66, no.~3, p. 1833–1859, 2024.

\bibitem{QianD2024}
D.~Qian, K.~Liu, J.~Wang, S.~Zhang, and X.~Yang, ``Attribute reduction based on
  directional semi-neighborhood rough set,'' \emph{International Journal of
  Machine Learning and Cybernetics}, 2024,
  https://doi.org/10.1007/s13042-024-02406-x.

\bibitem{FriedmanM1940}
M.~Friedman, ``A comparison of alternative tests of significance for the
  problem of $m$ rankings,'' \emph{The Annals of Mathematical Statistics},
  vol.~11, no.~1, pp. 86--92, 1940.

\bibitem{DemvsarJ2006}
J.~Dem{\v{s}}ar, ``Statistical comparisons of classifiers over multiple data
  sets,'' \emph{Journal of Machine Learning Research}, vol.~7, pp. 1--30, 2006.

\end{thebibliography}

\end{document}


\title{Supplementary Materials:\\ Margin-aware Fuzzy Rough Feature Selection: Bridging Uncertainty Characterization and Pattern Classification}

\author{Suping Xu, Lin Shang,~\IEEEmembership{Member,~IEEE,} Keyu Liu,\\
Hengrong Ju, Xibei Yang, and Witold Pedrycz,~\IEEEmembership{Life Fellow,~IEEE}

\thanks{S. Xu and W. Pedrycz are with the Department of Electrical and Computer Engineering, University of Alberta, Edmonton, AB T6G 2R3, Canada. (E-mail: supingxu@yahoo.com; suping2@ualberta.ca, wpedrycz@ualberta.ca)}
\thanks{L. Shang is with the Department of Computer Science and Technology, Nanjing University, Nanjing 210023, China, and also with the State Key Laboratory for Novel Software Technology, Nanjing University, Nanjing 210023, China. (E-mail: shanglin@nju.edu.cn)}
\thanks{K. Liu and X. Yang are with the School of Computer, Jiangsu University of Science and Technology, Zhenjiang 212003, China. (E-mail: kyliu@just.edu.cn, jsjxy\_yxb@just.edu.cn)}
\thanks{H. Ju is with the School of Artificial Intelligence and Computer Science, Nantong University, Nantong 226019, China. (E-mail: juhengrong@ntu.edu.cn)}

\thanks{Manuscript received X X, 2025; revised X X, 2025.}}

\markboth{IEEE TRANSACTIONS ON FUZZY SYSTEMS,~Vol.~X, No.~X, X~2025}%
{Shell \MakeLowercase{\textit{et al.}}: Bare Demo of IEEEtran.cls for IEEE Journals}

\maketitle

\subsection{Predictive Results}

\begin{figure*}[!t]
\centering
\includegraphics[width=0.9\textwidth]{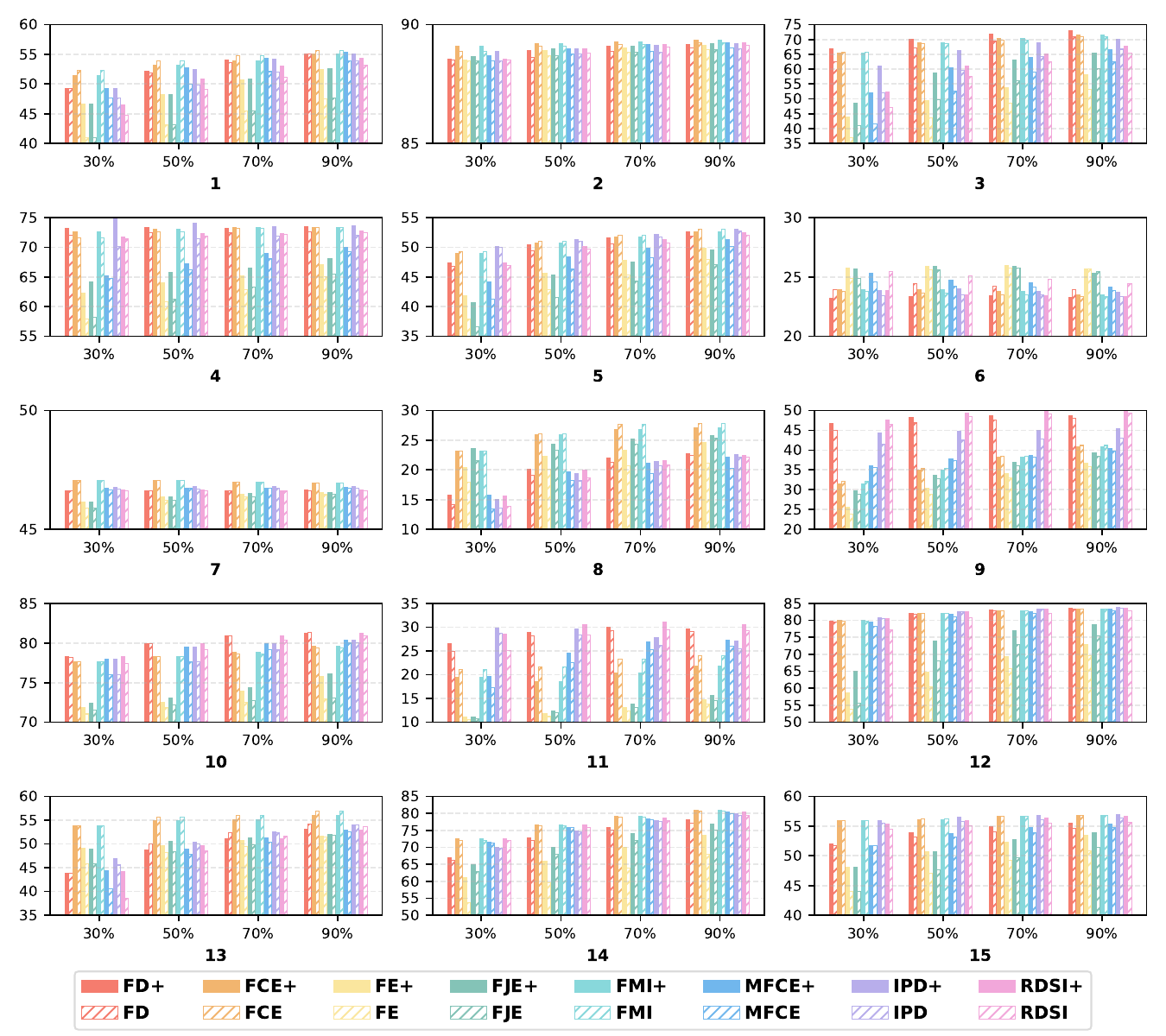}
\caption{Predictive performance (\texttt{SVM}) using the top 30\%, 50\%, 70\%, and 90\% of ranked features across $15$ datasets.}
\label{fig:8}
\end{figure*}

\begin{figure*}[!t]
\centering
\includegraphics[width=0.9\textwidth]{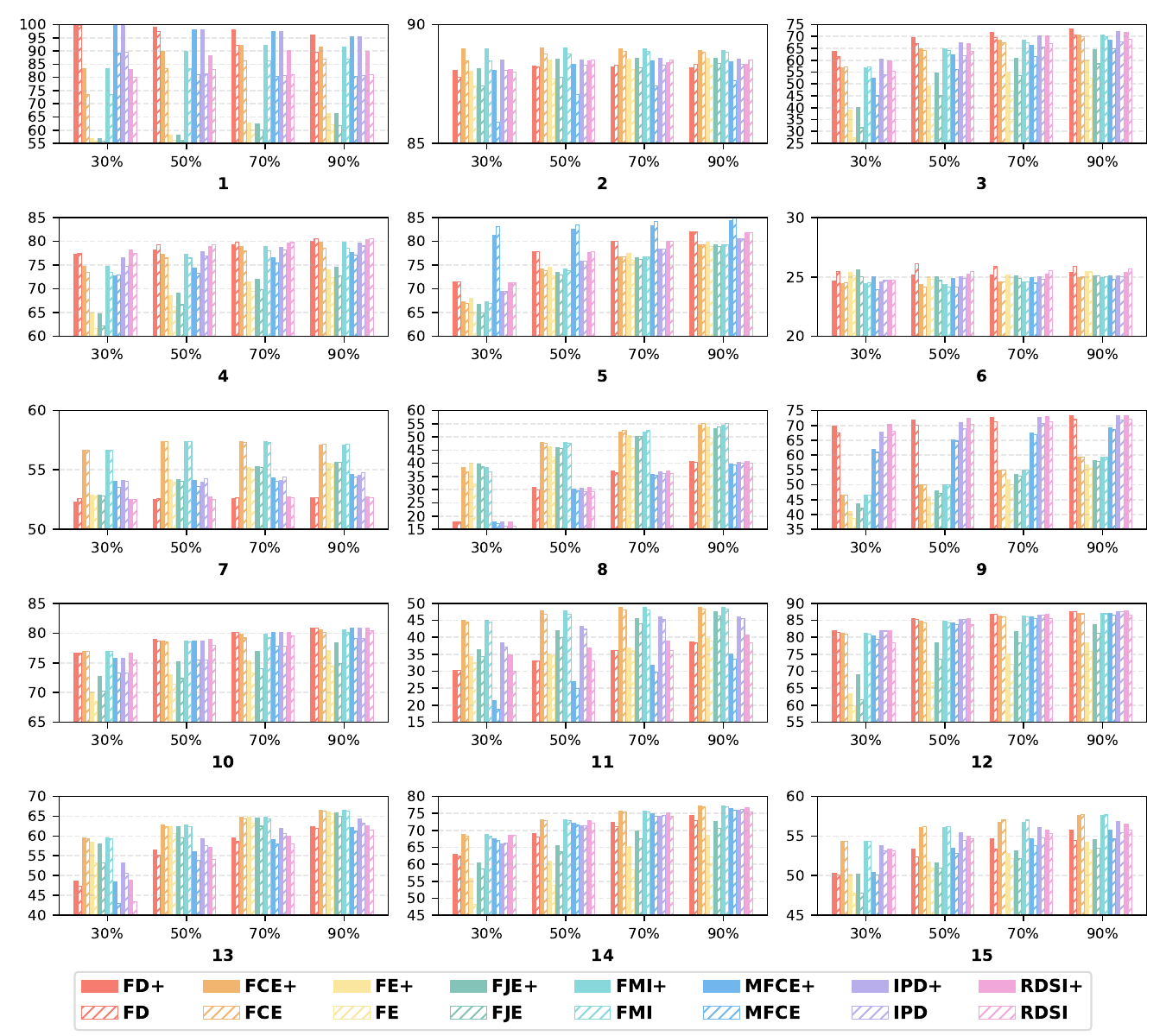}
\caption{Predictive performance (\texttt{KNN}) using the top 30\%, 50\%, 70\%, and 90\% of ranked features across $15$ datasets.}
\label{fig:9}
\end{figure*}

Figs.~\ref{fig:8} -- \ref{fig:9} presents detailed comparisons of predictive performance when employing the top 30\%, 50\%, 70\%, and 90\% of ranked features across $15$ datasets. The predictive performance in Figs.~\ref{fig:8} -- \ref{fig:9} are evaluated using the SVM and KNN classifiers, respectively. We can clearly observe that integrating each of the eight uncertainty measures, i.e., FD, FCE, FE, FJE, FMI, MFCE, IPD, and RDSI, into the MAFRFS framework leads to significant improvements compared to the FRFS framework in most cases.

Meanwhile, in Tabs.~\ref{tab:1} -- \ref{tab:3}, we present detailed predictive performance respectively using the top 30\%, 50\%, 70\%, and 90\% of ranked features. We can confirm that the MAFRFS framework delivers meaningful and reliable performance enhancements across different ratios of ranked features and diverse classifiers, highlighting its robust utility and broad applicability.

\begin{table*}[!t]
\caption{Predictive performance (\texttt{CART}) using the top 30\%, 50\%, 70\%, and 90\% of ranked features.}
\centering
\resizebox{0.95\textwidth}{!}{
\setlength{\tabcolsep}{2pt}
\begin{tabular}{c|c|cc|cc|cc|cc|cc|cc|cc|cc}
\hline
\textbf{ID} & \textbf{Rate} & \textbf{FD+} & \textbf{FD} & \textbf{FCE+} & \textbf{FCE} & \textbf{FE+} & \textbf{FE} & \textbf{FJE+} & \textbf{FJE} & \textbf{FMI+} & \textbf{FMI} & \textbf{MFCE+} & \textbf{MFCE} & \textbf{IPD+} & \textbf{IPD} & \textbf{RDSI+} & \textbf{RDSI}\\
\hline
\multirow{4}{*}{1} 
& 30\% & \textbf{99.776} & \textbf{99.776} & \textbf{83.679} & 74.424 & \textbf{71.879} & 70.515 & \textbf{71.879} & 70.515 & \textbf{83.679} & 74.424 & \textbf{99.776} & 99.487 & \textbf{99.776} & 98.342 & 96.457 & \textbf{96.831}\\
& 50\% & 99.770 & \textbf{99.811} & \textbf{90.167} & 84.460 & \textbf{79.263} & 78.987 & \textbf{79.263} & 78.987 & \textbf{90.167} & 84.460 & \textbf{99.355} & 97.308 & \textbf{99.355} & 96.609 & 97.362 & \textbf{97.457}\\
& 70\% & 99.431 & \textbf{99.438} & \textbf{92.823} & 88.618 & \textbf{84.027} & 83.597 & \textbf{83.826} & 83.597 & \textbf{92.823} & 88.618 & \textbf{99.223} & 97.643 & \textbf{99.223} & 97.142 & \textbf{97.798} & 97.670\\
& 90\% & \textbf{99.237} & 98.892 & \textbf{94.100} & 90.827 & \textbf{87.077} & 86.136 & \textbf{86.921} & 85.994 & \textbf{94.100} & 90.827 & \textbf{99.076} & 97.485 & \textbf{99.076} & 97.095 & \textbf{97.965} & 97.501\\
\hline                                                                                                   
\multirow{4}{*}{2} 
& 30\% & 86.225 & \textbf{86.601} & \textbf{88.531} & 87.895 & \textbf{85.036} & 84.296 & \textbf{84.954} & 84.296 & \textbf{88.531} & 87.895 & \textbf{88.719} & 88.376 & \textbf{88.858} & 88.470 & 86.546 & \textbf{86.906}\\
& 50\% & \textbf{86.590} & 86.535 & \textbf{87.851} & 87.533 & \textbf{85.650} & 85.053 & \textbf{85.631} & 85.208 & \textbf{87.851} & 87.533 & \textbf{88.468} & 88.059 & \textbf{88.639} & 88.296 & \textbf{87.000} & 86.991\\
\hline
\end{tabular}
}
\label{tab:1}
\end{table*}

\begin{table*}[!t]
\caption{Predictive performance (\texttt{SVM}) using the top 30\%, 50\%, 70\%, and 90\% of ranked features.}
\centering
\resizebox{0.95\textwidth}{!}{
\setlength{\tabcolsep}{2pt}
\begin{tabular}{c|c|cc|cc|cc|cc|cc|cc|cc|cc}
\hline
\textbf{ID} & \textbf{Rate} & \textbf{FD+} & \textbf{FD} & \textbf{FCE+} & \textbf{FCE} & \textbf{FE+} & \textbf{FE} & \textbf{FJE+} & \textbf{FJE} & \textbf{FMI+} & \textbf{FMI} & \textbf{MFCE+} & \textbf{MFCE} & \textbf{IPD+} & \textbf{IPD} & \textbf{RDSI+} & \textbf{RDSI}\\
\hline

\multirow{4}{*}{1} 
& 30\% & 49.208 & \textbf{49.252} & 51.482 & \textbf{52.309} & \textbf{46.680} & 41.009 & \textbf{46.680} & 41.009 & 51.482 & \textbf{52.309} & \textbf{49.208} & 47.633 & \textbf{49.208} & 47.576 & \textbf{46.512} & 44.726 \\                                                                                                                         
& 50\% & \textbf{52.145} & 51.910 & 53.099 & \textbf{53.949} & \textbf{48.259} & 43.095 & \textbf{48.259} & 43.095 & 53.099 & \textbf{53.949} & \textbf{52.663} & 49.906 & \textbf{52.490} & 49.871 & \textbf{50.872} & 49.077 \\                                                                                                                         
& 70\% & \textbf{53.981} & 53.535 & 53.960 & \textbf{54.804} & \textbf{50.643} & 45.511 & \textbf{50.827} & 45.511 & 53.960 & \textbf{54.804} & \textbf{54.351} & 52.080 & \textbf{54.122} & 52.056 & \textbf{53.071} & 51.163 \\                                                                                                                         
& 90\% & \textbf{55.015} & 54.991 & 54.999 & \textbf{55.655} & \textbf{52.370} & 47.559 & \textbf{52.513} & 47.686 & 54.999 & \textbf{55.655} & \textbf{55.303} & 53.869 & \textbf{55.125} & 53.850 & \textbf{54.307} & 53.156 \\
\hline
\end{tabular}
}
\label{tab:2}
\end{table*}

\begin{table*}[!t]
\caption{Predictive performance (\texttt{KNN}) using the top 30\%, 50\%, 70\%, and 90\% of ranked features.}
\centering
\resizebox{0.95\textwidth}{!}{
\setlength{\tabcolsep}{2pt}
\begin{tabular}{c|c|cc|cc|cc|cc|cc|cc|cc|cc}
\hline
\textbf{ID} & \textbf{Rate} & \textbf{FD+} & \textbf{FD} & \textbf{FCE+} & \textbf{FCE} & \textbf{FE+} & \textbf{FE} & \textbf{FJE+} & \textbf{FJE} & \textbf{FMI+} & \textbf{FMI} & \textbf{MFCE+} & \textbf{MFCE} & \textbf{IPD+} & \textbf{IPD} & \textbf{RDSI+} & \textbf{RDSI}\\
\hline                                                                                                   
\multirow{4}{*}{1} 
& 30\% & \textbf{99.767} & \textbf{99.767} & \textbf{83.368} & 73.488 & \textbf{56.903} & 55.694 & \textbf{56.903} & 55.694 & \textbf{83.368} & 73.488 & \textbf{99.767} & 89.086 & \textbf{99.767} & 89.556 & \textbf{82.991} & 79.893 \\
& 50\% & \textbf{99.165} & 97.239 & \textbf{89.975} & 83.169 & \textbf{58.265} & 56.279 & \textbf{58.265} & 56.279 & \textbf{89.975} & 83.169 & \textbf{98.152} & 81.103 & \textbf{98.152} & 81.385 & \textbf{88.086} & 82.964 \\
& 70\% & \textbf{98.059} & 92.000 & \textbf{91.992} & 86.343 & \textbf{62.609} & 59.963 & \textbf{62.442} & 59.963 & \textbf{91.992} & 86.343 & \textbf{97.336} & 80.474 & \textbf{97.336} & 80.676 & \textbf{90.146} & 81.012 \\
& 90\% & \textbf{96.108} & 89.438 & \textbf{91.389} & 86.995 & \textbf{66.332} & 62.075 & \textbf{66.202} & 61.800 & \textbf{91.389} & 86.995 & \textbf{95.545} & 80.474 & \textbf{95.545} & 80.630 & \textbf{89.953} & 80.892 \\
\hline                                                                                                           
\multirow{4}{*}{2}                                                                                               
& 30\% & \textbf{88.083} & 87.796 & \textbf{88.985} & 88.487 & \textbf{88.023} & 87.425 & \textbf{88.133} & 87.425 & \textbf{88.985} & 88.487 & \textbf{88.056} & 85.910 & \textbf{88.509} & 88.078 & \textbf{88.094} & 88.000 \\
\hline
\end{tabular}
}
\label{tab:3}
\end{table*}

\subsection{Parameter Sensitivity Analysis}

In the MAFRFS framework, the size of the pool ($sop$) is chosen from $\{2,3,4\}$. The  strategies employed for measuring the between-class margin -- whether global or local -- depend on the specific pattern classification tasks. Tab.~\ref{tab:4} presents the results of parameter sensitivity analysis conducted on FD+ and FCE+. The analysis explores the effects of varying the pool size ($sop = \{2,3,4\}$) and compares two strategies for measuring the between-class margin. From Tab.~\ref{tab:4}, we can conclude that: (i)  Algorithms within MAFRFS, when utilizing different uncertainty measures, frequently align with distinct optimal parameter combinations. (ii) For a given uncertainty measure, both global and local strategies occasionally yield the same feature subset, ultimately leading to identical predictive performance, regardless of the chosen value of $sop$. (iii) The pool size $sop$ plays a crucial role in effectively balancing the decrease of uncertainty in a pattern classification task and the increase of the margin between label classes.

\begin{table*}[!t]
\caption{Comparison of Predictive performance using \texttt{CART}: FD+ and FCE+ with $sop = \{2,3,4\}$ and two between-class margin strategies}
\centering
\resizebox{0.95\textwidth}{!}{
\setlength{\tabcolsep}{2pt}
\begin{tabular}{c|c|ccc|ccc|ccc|ccc}
\hline
\multirow{2}{*}{\textbf{ID}}  & \multirow{2}{*}{\textbf{Rate}} & \multicolumn{3}{c|}{\textbf{FD+: Global strategy}} & \multicolumn{3}{c|}{\textbf{FD+: Local strategy}} & \multicolumn{3}{c|}{\textbf{FCE+: Global strategy}} & \multicolumn{3}{c}{\textbf{FCE+: Local strategy}} \\
\cline{3-14}
& & $\mathbf{sop=2}$ & $\mathbf{sop=3}$ & $\mathbf{sop=4}$ & $\mathbf{sop=2}$ & $\mathbf{sop=3}$ & $\mathbf{sop=4}$ & $\mathbf{sop=2}$ & $\mathbf{sop=3}$ & $\mathbf{sop=4}$ & $\mathbf{sop=2}$ & $\mathbf{sop=3}$ & $\mathbf{sop=4}$ \\
\hline
\multirow{4}{*}{1} 
& 30\% & \textbf{99.776} & \textbf{99.776} & \textbf{99.776} & \textbf{99.776} & \textbf{99.776} & \textbf{99.776} & \textbf{83.679} & \textbf{83.679} & \textbf{83.679} & \textbf{83.679} & \textbf{83.679} & \textbf{83.679}\\ 
& 50\% & \textbf{99.770} & 99.644 & 99.644 & \textbf{99.770} & 99.644 & 99.644 & 90.154 & \textbf{90.167} & \textbf{90.167} & 90.154 & \textbf{90.167} & \textbf{90.167}\\ 
& 70\% & 99.256 & 99.107 & \textbf{99.431} & 99.256 & 99.107 & \textbf{99.431} & 92.814 & \textbf{92.823} & \textbf{92.823} & 92.814 & \textbf{92.823} & \textbf{92.823}\\ 
& 90\% & 98.947 & 98.986 & \textbf{99.237} & 98.947 & 98.986 & \textbf{99.237} & 94.093 & \textbf{94.100} & \textbf{94.100} & 94.093 & \textbf{94.100} & \textbf{94.100}\\
\hline
\multirow{4}{*}{2} 
& 30\% & \textbf{86.225} & 85.838 & 85.866 & \textbf{86.225} & 85.838 & 85.866 & \textbf{88.531} & 88.388 & 88.465 & \textbf{88.531} & 88.388 & 88.465\\ 
& 50\% & \textbf{86.590} & 85.954 & 86.411 & \textbf{86.590} & 85.954 & 86.411 & 87.688 & 87.721 & \textbf{87.851} & 87.688 & 87.721 & \textbf{87.851}\\ 
& 70\% & \textbf{86.857} & 86.238 & 86.546 & \textbf{86.857} & 86.238 & 86.546 & 87.356 & 87.386 & \textbf{87.477} & 87.356 & 87.386 & \textbf{87.477}\\ 
& 90\% & \textbf{86.956} & 86.397 & 86.632 & \textbf{86.956} & 86.397 & 86.632 & 87.266 & 87.267 & \textbf{87.351} & 87.266 & 87.267 & \textbf{87.351}\\ 
\hline
\multirow{4}{*}{3} 
& 30\% & 68.352 & 68.689 & \textbf{68.728} & 68.352 & 68.454 & 68.446 & 64.526 & \textbf{64.887} & \textbf{64.887} & 64.393 & 64.495 & 64.495\\ 
& 50\% & 72.794 & 73.368 & \textbf{73.532} & 72.822 & 72.963 & 73.048 & 69.243 & 69.440 & \textbf{69.610} & 68.627 & 69.200 & 69.323\\ 
& 70\% & 75.400 & \textbf{75.924} & 75.810 & 75.417 & 75.538 & 75.373 & 72.094 & 72.332 & \textbf{72.867} & 71.489 & 72.252 & 72.682\\ 
& 90\% & 76.908 & \textbf{77.253} & 77.098 & 76.913 & 76.939 & 76.759 & 74.213 & 74.404 & \textbf{74.796} & 73.777 & 74.422 & 74.663\\ 
\hline
\multirow{4}{*}{4} 
& 30\% & 72.810 & 73.024 & \textbf{73.798} & 72.810 & 73.024 & \textbf{73.798} & \textbf{71.982} & 71.661 & 71.635 & \textbf{71.982} & 71.661 & 71.635\\ 
& 50\% & 73.045 & \textbf{74.038} & 73.990 & 73.045 & \textbf{74.038} & 73.990 & \textbf{72.532} & 72.244 & 72.163 & \textbf{72.532} & 72.244 & 72.163\\ 
& 70\% & 72.859 & \textbf{74.279} & 73.294 & 72.859 & \textbf{74.279} & 73.294 & \textbf{72.505} & 72.047 & 71.978 & \textbf{72.505} & 72.047 & 71.978\\ 
& 90\% & 72.071 & \textbf{73.406} & 72.320 & 72.071 & \textbf{73.406} & 72.320 & \textbf{72.115} & 71.741 & 71.750 & \textbf{72.115} & 71.741 & 71.750\\ 
\hline
\multirow{4}{*}{5} 
& 30\% & 68.720 & \textbf{68.864} & 68.339 & 68.720 & \textbf{68.864} & 68.339 & 79.364 & 79.408 & 79.388 & 79.384 & \textbf{79.428} & 79.408\\ 
& 50\% & \textbf{75.503} & 75.378 & 74.811 & \textbf{75.503} & 75.378 & 74.811 & 83.384 & 83.409 & 83.301 & 83.397 & \textbf{83.428} & 83.311\\ 
& 70\% & \textbf{77.819} & 77.717 & 77.263 & \textbf{77.819} & 77.717 & 77.263 & 84.653 & 84.600 & 84.522 & \textbf{84.662} & 84.641 & 84.552\\ 
& 90\% & 80.025 & \textbf{80.134} & 79.915 & 80.025 & \textbf{80.134} & 79.919 & 85.789 & 85.750 & 85.696 & 85.793 & \textbf{85.798} & 85.736\\ 
\hline
\multirow{4}{*}{6} 
& 30\% & 24.170 & \textbf{24.531} & 24.287 & 24.106 & \textbf{24.531} & 24.224 & 25.054 & 25.280 & 25.433 & \textbf{25.469} & 24.892 & 25.054\\ 
& 50\% & \textbf{24.889} & 24.807 & 24.740 & 24.621 & 24.781 & 24.559 & 24.683 & \textbf{25.260} & 25.080 & 25.142 & 24.863 & 24.832\\ 
& 70\% & \textbf{25.244} & 24.724 & 25.028 & 25.077 & 24.515 & 24.895 & 24.788 & \textbf{25.590} & 25.335 & 25.301 & 25.286 & 25.119\\ 
& 90\% & \textbf{25.415} & 24.630 & 24.982 & 25.355 & 24.468 & 24.930 & 24.786 & \textbf{25.453} & 25.308 & 25.256 & 25.216 & 25.164\\ 
\hline
\multirow{4}{*}{7} 
& 30\% & \textbf{54.549} & 53.243 & 52.998 & \textbf{54.549} & 53.243 & 52.998 & \textbf{58.881} & 58.218 & 58.383 & \textbf{58.881} & 58.218 & 58.383\\ 
& 50\% & \textbf{54.625} & 53.884 & 53.272 & \textbf{54.625} & 53.884 & 53.272 & \textbf{59.700} & 59.058 & 59.239 & \textbf{59.700} & 59.058 & 59.239\\ 
& 70\% & \textbf{54.194} & 54.015 & 53.378 & \textbf{54.194} & 54.015 & 53.378 & \textbf{59.916} & 59.384 & 59.708 & \textbf{59.916} & 59.384 & 59.708\\ 
& 90\% & \textbf{54.352} & 54.130 & 53.795 & \textbf{54.352} & 54.130 & 53.795 & \textbf{60.001} & 59.620 & 59.997 & \textbf{60.001} & 59.620 & 59.997\\ 
\hline
\multirow{4}{*}{8} 
& 30\% & 24.706 & \textbf{25.147} & 25.074 & 24.118 & \textbf{25.147} & 25.074 & 35.147 & 35.147 & 35.147 & \textbf{35.441} & \textbf{35.441} & \textbf{35.441}\\ 
& 50\% & 37.773 & \textbf{38.403} & 37.815 & 37.437 & \textbf{38.403} & 37.815 & 43.067 & 42.941 & 42.395 & \textbf{43.403} & 43.277 & 42.773\\ 
& 70\% & 44.324 & \textbf{44.824} & 44.382 & 44.176 & \textbf{44.824} & 43.971 & 46.059 & 46.118 & 46.059 & 46.382 & 46.353 & \textbf{46.412}\\ 
& 90\% & 47.308 & \textbf{47.624} & 47.308 & 47.262 & \textbf{47.624} & 47.081 & 48.756 & 48.688 & 48.462 & \textbf{49.005} & 48.891 & 48.733\\ 
\hline
\multirow{4}{*}{9} 
& 30\% & 55.144 & 57.623 & 56.399 & 56.142 & \textbf{58.323} & 57.479 & 45.545 & \textbf{45.576} & 45.123 & 45.432 & 45.504 & 45.021\\ 
& 50\% & 57.407 & 59.574 & 58.315 & 57.963 & \textbf{60.037} & 59.568 & 48.827 & \textbf{48.895} & 48.704 & 48.753 & 48.877 & 48.673\\ 
& 70\% & 59.229 & 60.699 & 59.803 & 59.435 & \textbf{61.685} & 60.533 & 51.380 & 51.290 & 50.896 & \textbf{51.478} & 51.380 & 51.071\\ 
& 90\% & 60.528 & 61.444 & 60.669 & 60.816 & \textbf{62.353} & 61.252 & 54.468 & 54.424 & 54.002 & \textbf{54.523} & 54.462 & 54.256\\ 
\hline
\multirow{4}{*}{10} 
& 30\% & \textbf{73.302} & 73.284 & 73.284 & \textbf{73.302} & 73.284 & 73.284 & \textbf{73.661} & \textbf{73.661} & \textbf{73.661} & \textbf{73.661} & \textbf{73.661} & \textbf{73.661}\\ 
& 50\% & \textbf{75.788} & 75.701 & 75.396 & \textbf{75.788} & 75.701 & 75.396 & \textbf{75.760} & \textbf{75.760} & \textbf{75.760} & \textbf{75.760} & \textbf{75.760} & \textbf{75.760}\\ 
& 70\% & \textbf{77.191} & 76.753 & 76.672 & \textbf{77.191} & 76.753 & 76.672 & \textbf{76.999} & \textbf{76.999} & \textbf{76.999} & \textbf{76.999} & \textbf{76.999} & \textbf{76.999}\\ 
& 90\% & \textbf{78.133} & 77.931 & 77.862 & \textbf{78.133} & 77.931 & 77.862 & 78.086 & \textbf{78.125} & \textbf{78.125} & 78.086 & \textbf{78.125} & \textbf{78.125}\\ 
\hline
\multirow{4}{*}{11} 
& 30\% & 31.076 & \textbf{31.144} & 30.621 & 31.041 & 30.886 & 30.330 & 51.512 & \textbf{52.221} & 51.924 & 51.494 & 51.974 & 51.714\\ 
& 50\% & 34.062 & \textbf{34.199} & 33.794 & 34.042 & 34.023 & 33.609 & 56.718 & \textbf{57.134} & 56.914 & 56.702 & 56.991 & 56.823\\ 
& 70\% & \textbf{36.646} & 36.643 & 36.530 & 36.604 & 36.524 & 36.388 & 59.065 & \textbf{59.382} & 59.178 & 59.051 & 59.285 & 59.142\\ 
& 90\% & 41.317 & \textbf{41.401} & 41.347 & 41.285 & 41.313 & 41.228 & 60.201 & \textbf{60.412} & 60.244 & 60.190 & 60.342 & 60.219\\
\hline 
\multirow{4}{*}{12} 
& 30\% & 80.176 & 80.199 & \textbf{80.239} & 80.183 & 80.208 & 80.210 & 79.936 & 80.034 & \textbf{80.211} & 79.963 & 79.992 & 80.020\\ 
& 50\% & 82.381 & 82.358 & 82.346 & \textbf{82.416} & 82.378 & 82.309 & 81.939 & 82.017 & \textbf{82.128} & 81.609 & 81.464 & 81.397\\ 
& 70\% & 83.216 & 83.308 & \textbf{83.324} & 83.231 & 83.276 & 83.280 & 82.899 & 82.958 & \textbf{83.041} & 82.692 & 82.552 & 82.540\\ 
& 90\% & 83.799 & 83.919 & \textbf{83.938} & 83.804 & 83.902 & 83.914 & 83.560 & 83.601 & \textbf{83.692} & 83.399 & 83.283 & 83.294\\ 
\hline
\multirow{4}{*}{13} 
& 30\% & 49.433 & 50.000 & 49.716 & 49.433 & 49.693 & \textbf{50.024} & 60.000 & 60.000 & \textbf{60.095} & 60.000 & 60.000 & 60.000\\ 
& 50\% & \textbf{57.210} & 56.961 & 56.488 & \textbf{57.210} & 56.777 & 57.092 & 62.464 & 63.055 & 62.832 & 62.464 & 63.055 & \textbf{63.186}\\ 
& 70\% & \textbf{60.234} & 59.673 & 58.737 & \textbf{60.234} & 59.496 & 59.200 & 64.125 & \textbf{64.657} & 64.027 & 64.125 & \textbf{64.657} & 64.273\\ 
& 90\% & \textbf{62.840} & 62.345 & 61.488 & \textbf{62.840} & 62.212 & 61.924 & 65.655 & 66.046 & 65.610 & 65.603 & \textbf{66.105} & 65.669\\ 
\hline
\multirow{4}{*}{14} 
& 30\% & 59.530 & \textbf{59.793} & 59.643 & 59.530 & 59.723 & 59.617 & \textbf{66.053} & 65.970 & 65.970 & 65.237 & 65.177 & 65.177\\ 
& 50\% & 64.708 & 65.186 & \textbf{65.368} & 64.708 & 65.140 & 65.294 & \textbf{69.514} & 69.396 & 69.396 & 68.992 & 68.938 & 68.938\\ 
& 70\% & 67.387 & 68.067 & \textbf{68.151} & 67.387 & 68.037 & 68.090 & \textbf{71.401} & 71.306 & 71.306 & 71.006 & 70.973 & 70.973\\ 
& 90\% & 69.073 & 69.694 & \textbf{69.747} & 69.073 & 69.673 & 69.694 & \textbf{72.376} & 72.304 & 72.304 & 72.067 & 72.049 & 72.049\\
\hline 
\multirow{4}{*}{15} 
& 30\% & 52.220 & \textbf{52.491} & 52.304 & 52.053 & \textbf{52.491} & 52.304 & 55.931 & 55.931 & \textbf{56.056} & 55.931 & 55.931 & \textbf{56.056}\\ 
& 50\% & 53.671 & 54.472 & \textbf{54.634} & 53.896 & 54.384 & \textbf{54.634} & 56.886 & 56.785 & 57.048 & 56.886 & 56.785 & \textbf{57.111}\\ 
& 70\% & 55.025 & 56.383 & \textbf{56.446} & 55.302 & 56.339 & \textbf{56.446} & 57.634 & 57.715 & 57.706 & 57.634 & 57.634 & \textbf{57.813}\\ 
& 90\% & 55.993 & 57.216 & \textbf{57.327} & 56.382 & 57.189 & 57.300 & 58.189 & 58.259 & 58.252 & 58.085 & 58.168 & \textbf{58.328}\\ 
\hline 
\end{tabular}
}
\label{tab:4}
\end{table*}

\vfill